\documentclass[conference]{IEEEtran}

\IEEEoverridecommandlockouts                              % This command is only needed if 
\usepackage{cite}
\def\BibTeX{{\rm B\kern-.05em{\sc i\kern-.025em b}\kern-.08em
    T\kern-.1667em\lower.7ex\hbox{E}\kern-.125emX}}
    
\usepackage{amsmath}
\usepackage{bm}
\usepackage{amsthm}
\usepackage{amssymb}
\usepackage{amsfonts}
\usepackage{float}
\usepackage{soul}
\usepackage{algorithmic}
\usepackage{tensor}
\usepackage{graphicx}
\usepackage{physics}
\usepackage{textcomp}
\usepackage{wrapfig}
\usepackage{caption}
\usepackage{hyperref}
\usepackage{subcaption}
\usepackage{multirow}
\usepackage{comment}
\usepackage{steinmetz}
\usepackage{mathtools}
\usepackage{cuted}

\usepackage[table]{xcolor}
\theoremstyle{definition}
\newtheorem{remark}{Remark}
\title{\LARGE \bf
Analysis of the Effect of Time Delay for Unmanned Aerial Vehicles with Applications to Vision Based Navigation}

\author{
        Muhammad~Ahmed~Humais\href{https://orcid.org/0000-0001-6237-6394}{\includegraphics[scale=0.75]{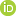}},
        Mohamad~Chehadeh\href{https://orcid.org/0000-0002-9430-3349}{\includegraphics[scale=0.75]{figures/orcid.png}},~\IEEEmembership{Member,~IEEE,}\\
        Igor~Boiko\href{https://orcid.org/0000-0003-4978-614X}{\includegraphics[scale=0.75]{figures/orcid.png}},~\IEEEmembership{Senior~Member,~IEEE}
        and~Yahya~Zweiri\href{https://orcid.org/0000-0003-4331-7254}{\includegraphics[scale=0.75]{figures/orcid.png}},~\IEEEmembership{Member,~IEEE}% <-this % stops a space
\thanks{This work was supported by Khalifa University Grant CIRA-2020-082.

M. Chehadeh, M. Humais, I. Boiko, and Y. Zweiri are with the Center for Autonomous Robotic Systems, Khalifa University, Abu Dhabi, United Arab Emirates. Also, I. Boiko is with the Department of Electrical Engineering and Computer Science, and Y. Zweiri with the Department of Aerospace Engineering, both at Khalifa University, Abu Dhabi, United Arab Emirates.}

\thanks{*Muhammad Humais and Mohamad Chehadeh contributed equally to this work.}% <-this % stops a space
}
\begin{document}

\maketitle
\thispagestyle{empty}
\pagestyle{empty}

%%%%%%%%%%%%%%%%%%%%%%%%%%%%%%%%%%%%%%%%%%%%%%%%%%%%%%%%%%%%%%%%%%%%%%%%%%%%%%%%
\begin{abstract}
In this paper, we analyze the effect of time delay dynamics on controller design for Unmanned Aerial Vehicles (UAVs) with vision based navigation. Time delay is an inevitable phenomenon in cyber-physical systems, and has important implications on controller design and trajectory generation for UAVs. The impact of time delay on UAV dynamics increases with the use of the slower vision based navigation stack. We show that the existing models in the literature, which exclude time delay, are unsuitable for controller tuning since a trivial solution for minimizing an error cost functional always exists. The trivial solution that we identify suggests use of infinite controller gains to achieve optimal performance, which contradicts practical findings. We avoid such shortcomings by introducing a novel nonlinear time delay model for UAVs, and then obtain a set of linear decoupled models corresponding to each of the UAV control loops. The cost functional of the linearized time delay model of angular and altitude dynamics is analyzed, and in contrast to the delay-free models, we show the existence of finite optimal controller parameters. Due to the use of time delay models, we experimentally show that the proposed model accurately represents system stability limits. Due to time delay consideration, we achieved a tracking results of RMSE 5.01 cm when tracking a lemniscate trajectory with a peak velocity of 2.09 m/s using visual odometry (VO) based UAV navigation, which is on par with the state-of-the-art.
\end{abstract}

\section{INTRODUCTION}
Unmanned Aerial Vehicles (UAVs) became very popular in recent years for a wide range of applications such as inspection \cite{inspection2013}, agriculture \cite{agriculture}, entertainment \cite{entertainment}, package delivery \cite{drone_delivery}, etc. Accurate trajectory tracking is required for many of these applications and is very critical for safe autonomous operation of the UAVs. For operations in GPS denied environments, vision based localization using on-board camera is the most attractive solution, and has been implemented successfully in recent works \cite{faessler2016autonomous, mohta2018fast}. However the vision feedback introduces additional dynamics to overall UAV control system and affects the trajectory tracking performance as well as the stability of the platform.
% Accurate trajectory tracking may be required for many of these applications such as aerial surveillance, entertainment, package delivery, inspection, etc \cite{inspection2013},\cite{transportation}.
% For many of these applications, accurate trajectory tracking is required for the success of the mission. Applications that require navigation in cluttered environments such as last mile package delivery %Be specific, and cite specifically

% Rewrite and integrate well to avoid repetition

% Trajectory tracking requires hardware, sensors, perception algorithms, and control to make them work in synergy
\subsection{Related Work}
Multi-rotor UAVs are fairly complex non-linear and under-actuated systems which are inherently unstable. Hence, the control of UAV is a challenging task and has been addressed widely in the robotics research community \cite{uav_Control_survey}. The most common control approach uses differential flatness for feedback linearization, combined with a hierarchical controller structure with fixed controller gains \cite{lee2010geometric,mahony2012multirotor,mellinger2012trajectory,PhD18_Faessler}. Other popular control approaches are incremental nonlinear dynamic inversion (INDI) \cite{smeur2018cascaded}, model predictive control (MPC) \cite{mpc}, Linear Quadratic Regulator (LQR) \cite{lqr2015} control, non-linear controllers accounting for non-linear drag \cite{nonlinear} and many others. These different controller designs are effective for stable flight but for accurate trajectory tracking these controllers need to be carefully tuned in order to achieve performance goals. However, tuning these controllers require accurate dynamic model of the UAV or extensive manual tuning, which is much more complex than tuning of PID controllers due to absence of well developed methods. Overall, the controller design and tuning for trajectory tracking has been extensively studied in the literature\cite{trackingsurvey}. One way to tune the controllers is to repeat the trajectory and iteratively learn to adapt the control inputs. Many different solutions based on this idea have been proposed \cite{ILC_1}, \cite{ILC_2}. A Deep Neural Network (DNN) based approach has also been proposed to learn inverse models and use it to adapt the control input for trajectory tracking \cite{trajectory_dnn}. These approaches have mainly two drawbacks, the controllers are tuned on specific trajectories and it requires large amount of experimental data for best performance. Hence, considering an accurate dynamic model could be more advantageous and once the model is identified the controller can be tuned for tracking any feasible trajectory. Since the accurate dynamic model of the UAV is usually not available and the model mismatch severely impacts the control performance, \cite{vertical_window} performs several experimental trials to tune the controllers for achieving precise and aggressive maneuvers. In multiple recent works \cite{faessler2017differential,Jia2022wind}, the authors estimated the rotor drag by repeating a predefined trajectory and used it to boost the performance of the designed controllers. However, they also needed multiple flights on predefined trajectory to collect data and compute drag coefficients. Further, they also compensate for the delays but the method for estimating delays is unclear. Similarly, the work of \cite{carlos2020efficient} used a state prediction approach to compensate for communication delays using nonlinear MPC.

% (even they compensate for delays manually). (Present challenge, solution, drawback, solution, drawback, etc.) 

% (Common flaw - addressed by our recent work - but still it is not suitable for gps-denied. Importance of GPS denied - State the gap clearly: performance limits are not known where I should spend my time own, automation of tuning)
Most of these works for accurate and agile trajectory tracking have been done under external localization systems (e.g. motion capture) \cite{vertical_window, diff_flat,Ayyad2020, trajectory_dnn}. Such systems are only appropriate for development and evaluation purposes. These systems are expensive and require pre-installation in the environment and further limit the operation area hence not suitable for many applications. The best alternative is to use on-board vision sensors as they are much cheaper than 3D laser scanners and light-weight, however vision-based localization requires huge computational resources. Recent development of efficient and fast visual odometry algorithms such as PTAM \cite{ptam, ptam2}, SVO \cite{svo}, etc. has enabled the use of on-board vision-based localization for autonomous navigation of UAVs \cite{faessler2016autonomous, mohta2018fast, special_issue}. The use of vision sensor and visual odometry (VO) pipeline introduces certain delays and dynamics to the overall UAV control system. If not included in the dynamic model of UAV it immensely affects trajectory tracking performance and can cause instability in worst case. However, this effect is not studied in the literature, mainly because the use of on-board vision for positional tracking of UAVs has become viable only recently. 
% Hence there is a need to systematically study the effect of vision-based positional feedback in the UAV control system and account for that in the controller tuning.

As can be seen from the above survey, the effect of time delay is widely acknowledged but was not addressed methodologically. The recent work of \cite{sharma2021control} proposed a nonlinear time delay model with state predictors to compensate for the time delay. Such analysis is not of much interest for practical applications, since it is assumed that the time delay and UAV parameters are known a priori, and hence \cite{sharma2021control} was limited to simulations. Another work by \cite{daly2015coordinated} made use of simpler quadrotor models with time delay to analyze stability of UAV controllers for the task of landing on a moving ground vehicle. Another experimental work by \cite{liu2019robust} demonstrated the use of robust control methods to account for time delays in the UAV platform. The gains of the controllers of \cite{daly2015coordinated} and \cite{liu2019robust} were tuned manually based on flight tests. The work of \cite{poksawat2016automatic} attempted to avoid the manual tuning effort using relay feedback test (RFT) for the attitude loops of a fixed wing UAV while accounting for time delay dynamics. The limitation of \cite{poksawat2016automatic} is that the UAV was tuned on a bench setup. The work of \cite{chehadeh2019design} overcame this limitation through the use of the modified relay feedback test (MRFT), which allows one to excite stable test oscillations even in the open-loop unstable dynamics, and performed gains tuning online for attitude loops of a quadrotor UAV.

To overcome the noted limitations  arising from the complex dynamics and non-negligible delay in the control loop, we recently proposed a novel framework \cite{Ayyad2020} based on deep neural networks and the modified relay feedback test (DNN-MRFT), which can estimate the delays and the dynamics of the linearized decoupled UAV system and provides near optimal controllers for trajectory tracking in real time. The results in \cite{ayyad2021tcst} extended the DNN-MRFT application to the underactuated lateral dynamics. Yet the literature is short of a multirotor UAV model that captures nonlinear dynamic models similar to those in \cite{faessler2017differential}, while accounting for time delays and actuator dynamics. Such models can be used to derive valid linearized models suitable for controller design.

\subsection{Contributions}
Based on the above, we present a novel nonlinear time delay model suitable for high performance trajectory tracking tasks of multirotor UAVs. The main challenge that motivates such development is the sensor delays and high computational overhead coming from vision-based localization running on a resource constrained on-board computer. Hence, the contributions of the paper stem from the analysis of the nonlinear time delay dynamics, and can be summarized as follows:
\begin{enumerate}
    \item We propose a full nonlinear model suitable for vision based multirotor UAVs that accounts for time delay and actuator dynamics. The nonlinear models widely used in literature neglect delay dynamics \cite{faessler2017differential,lee2013nonlinear,mellinger2012trajectory,mohta2018fast}. The nonlinear model is then linearized and decoupled to obtain linear time delay models suitable for identification and controller tuning.
    \item We prove using analytical cost functionals that the delay free models cannot be used for controller tuning due to the presence of a trivial global minimum that happens when controller gains are infinite.
    \item Using the developed time delay model with the DNN-MRFT approach for identification and tuning, we experimentally achieve state-of-the-art trajectory tracking results of UAVs with vision based navigation. 
\end{enumerate}

The consideration of time delay in the model allowed us to accurately tune the feedback controllers in real-time, and consider the delay dynamics in trajectory generation. As a result, we were able to achieve a tracking performance of RMSE 5.01 cm and ATE 2.7 cm when tracking a lemniscate trajectory with a peak velocity of 2.09 m/s. These results were achieved using visual odometry (VO) based UAV navigation and are on par with state-of-the-art results while having the advantage of being fully tuned in real-time. We also demonstrate the adequacy of the considered model as it accurately predicts stability margins which verified experimentally. A supplementary video for this work showing the experiments can be found at https://youtu.be/XdLdL9QeSWE.

\subsection{Outline}
This paper is organized as follows. First the nonlinear time delay model and the linearized and decoupled models of multirotor UAVs are presented in Section \ref{sec_model_control}. Then an evaluation of the suitability of delay-free models and time delay models for controller tuning based on analytical cost functionals is presented in Section \ref{sec_eval_suitability_cost_func}. Then in Section \ref{sec:dnnmrft} we show how DNN-MRFT can be used for the identification of the model dynamic parameters, including time delay, followed by the tuning of the controller parameters. Finally, Section \ref{sec_exp_results} shows the experimental results obtained.

\section{Modeling and Control}
\label{sec_model_control}
The model parameters identification and the controller design of the closed loop system using DNN-MRFT requires linear SISO models. The UAV model is nonlinear for the following two reasons: 1) the angular kinematics, and 2) the complex aerodynamic and gyroscopic kinetics. We use the feedback linearization approach suggested in \cite{Lee2010,Mellinger2011} to linearize the angular kinematics. We account for the complex aerodynamic and gyroscopic kinetics by the use of a first order drag model that gets estimated using online identification.

In this section, we show the steps followed to obtain the required linear SISO models. We first present the nonlinear UAV model with time delay. Then we present the feedback linearization law and the resulting closed loop system dynamics. Lastly, the dynamics are decoupled to obtain six SISO models for each DoF.

\subsection{Reference Frames and Conventions}
We define an inertial frame \(\mathcal{F}_I\) having basis \( \bm{[i_x, i_y, i_z]} \) with \(\bm{i_z}\) antiparallel to the gravity vector, and a body-fixed reference frame \(\mathcal{F}_B\) centered at center of gravity of the UAV with rotation matrix \(  {}^I_B\bm{R} = \bm{[b_x, b_y, b_z]} \in \text{SO(3)} \), which gives the transformation from \(\mathcal{F}_B\) to \(\mathcal{F}_I\), where \( \bm{b_z} \) is parallel to the thrust vector. We also define the horizon frame $\mathcal{F}_H$ with its origin is coincident with the origin of $\mathcal{F}_I$, its basis \(\bm{h_z}\) being coincident with \(\bm{i_z}\), and it is yaw aligned with $\mathcal{F}_B$. A vector can be expressed in a particular reference frame, e.g. \({}^I\bm{p}\) is the position vector expressed in the inertial frame. The components of a vector are referred to with the subscripts as in \({}^I\bm{p}=[{}^Ip_x {}^Ip_y {}^Ip_z]^T\). For compatibility of notation with vector quantities we use \(K_x\) to represent the element \(K_{11}\) in a diagonal matrix, and so on.

\subsection{Time Delay UAV Model}
\label{sec_time_delay_model}
We define the motor commands as follows:
\begin{equation}
    \begin{bmatrix}
    \bm{u_\eta} \\
    u_T
    \end{bmatrix} = \bar{G}\bm{u_i}
\end{equation}
where \(\bm{u_\eta}=[u_{b_x} u_{b_y} u_{b_z}]^T\) represents torque commands around \(\mathcal{F}_B\) bases, \(u_T\) is the thrust command, and \(\bm{u_i}\in[0,1]\) is the dimensionless individual motor command with \(i\in \{1,...,\mu_n\}\) where \(\mu_n\) represents the number of propellers used. \(\bar{G}\in\mathbb{R}^{4\times \mu_n}\) provides a static map independent of UAV dynamics, with \(\text{rank}(\bar{G})=4\) and its Moore–Penrose inverse \(\bar{G}^{+}\) is defined and unique. The individual propulsion system thrust and moment dynamics are given by:
\begin{equation}
\label{eq_act_dynamics_time}
\begin{aligned}
    F_i(t) &= k_F u_i(t-\tau_p)-T_p \dot{F}_i(t) \\
    M_i(t) &= k_M u_i(t-\tau_p)-T_p \dot{M}_i(t)
\end{aligned}
\end{equation}
where \(k_F\), \(k_M\), \(\tau_p\), and \(T_p\) are the thrust gain, moment gain, propulsion system time delay, and propulsion system time constant respectively. Note that we assume that all propulsion units are matched, i.e. the parameters \(k_F\), \(k_M\), \(\tau_p\), and \(T_p\) are the same for all rotors. Also, it is assumed that the thrust and moments applied to the rigid body are defined by the relation:
\begin{equation}
    \begin{bmatrix}
    \bm{M} \\
    F
    \end{bmatrix} = G_F\bm{F_p}+G_M\bm{M_p}
\end{equation}
where \(\bm{F_p}=[F_1 F_2 ... F_{\mu_n}]^T\) and \(\bm{M_p}=[M_1 M_2 ... M_{\mu_n}]^T\). \(G_F\) and \(G_M\) are static maps which may contain UAV dynamic parameters. For example, for a quadrotor UAV they are defined as:
\begin{equation}
\begin{aligned}
    G_F &= \begin{bmatrix}
    l_y/\sqrt{2} & -l_y/\sqrt{2} & -l_y/\sqrt{2} & l_y/\sqrt{2} \\
    l_x/\sqrt{2} & l_x/\sqrt{2} & -l_x/\sqrt{2} & -l_x/\sqrt{2} \\
    0 & 0 & 0 & 0 \\
    1 & 1 & 1 & 1
    \end{bmatrix} \\
    G_M &= \begin{bmatrix}
    0 & 0 & 0 & 0 \\
    0 & 0 & 0 & 0 \\
    1 & -1 & 1 & -1 \\
    0 & 0 & 0 & 0
    \end{bmatrix}
\end{aligned}
\end{equation}

The UAV dynamics are then given by:
\begin{equation}
\label{eq_uav_dynamics}
    \begin{aligned}
        {}^I\dot{\bm{p}}&={}^I\bm{v}\\
        {}^I\dot{\bm{v}}&=-g\bm{i_z}+\frac{F}{m}\bm{b_z}-{}_B^I{R}D{}_I^B{R}{}^I\bm{v}\\
        \dot{R}&={}_B^IR{}^B\bm{\omega}\\
        \bm{\dot{\omega}}&=J^{-1}(\bm{M}-\bm{\omega}\times J\bm{\omega}-\bm{M_g}-A{}_I^B{R}{}^I{\bm{v}}-B\bm{\omega})
    \end{aligned}
\end{equation}
where the diagonal matrices \(D\), \(J\), \(A\) and \(B\) represents profile drag and inflow motion drag due to translational motion, moment of inertia, drag due to blade flapping, and rotational drag due to body profile and inflow motion, respectively. The vector \(\bm{M_g}\) represents gyroscopic moments due to the interaction between rotating propellers and rotating UAV body.

\subsection{Feedback Linearization and Closed Loop Control}
The proposed feedback controllers minimize four errors: \(\bm{e_p}={}^I\bm{p_r}(t)-{}^I\bm{p}(t-\tau_{pm})\), \(\bm{{e}_v}={}^I\bm{v_r}(t)-{}^I\bm{v}(t-\tau_{vm})\), \(\bm{e_\eta}={}^I\bm{\eta_r}(t)-{}^I\bm{\eta}(t-\tau_{\eta m})\), and \(\bm{e_\omega}=-{}^I\bm{\omega}(t-\tau_{\omega m})\) corresponding to the errors in position, velocity, attitude, and attitude rate respectively (note that we use minimum snap trajectory generation to get the reference signals \cite{mellinger2011minimum}). It is reasonable to assume that \(\tau_{pm}=\tau_{vm}\) and \(\tau_{\eta m}=\tau_{\omega m}\) as each of the pairs is usually obtained from the same measurement source. The time delay terms account for the delay in measurements such that the best possible available estimate is delay, e.g. \({}^I\bm{\hat{p}}(t)={}^I\bm{p}(t-\tau_{pm})\). The position controller is then defined as:
\begin{equation}
\label{eq_outer_loop_ctrl}
\begin{aligned}
    {}^I\bm{a_d}=&{}_H^IR(K_p{}_I^HR\bm{e_p}+K_v{}_I^HR\bm{{e}_v}+K_i{}_I^HR\bm{\bar{e}_p})\\
    &+{}^I\bm{a_r}+\hat{a}_g\bm{i_z}
\end{aligned}
\end{equation}
where \(K_p\), \(K_v\), \(K_i\) are diagonal constant matrices, which have the meaning of the proportional, derivative and integral gain, respectively. \(\hat{a}_g\) is the estimated acceleration due to gravity, and \(\bm{\bar{e}_p}\) is an augmented controller state given by:
\begin{equation}
    \bm{\bar{e}_p}=\int_0^t\bm{e_p}dt
\end{equation}

The reference attitude \({}^I\bm{\eta_r}\) can be then calculated as:
\begin{equation}
\label{eq_fb_lin_r_bc}
    \begin{aligned}
    \bm{c_z}&=\frac{{}^I\bm{a_d}}{\|{}^I\bm{a_d}\|}\\
    \bm{c_y}&=\frac{\bm{c_z} \times [\cos(\psi_{r}) \; \sin(\psi_{r}) \; 0]^T}{\| \bm{c_z} \times [\cos(\psi_{r}) \; \sin(\psi_{r}) \; 0]^T \|}\\
    \bm{c_x} &= \bm{c_y} \times \bm{c_z}
    \end{aligned}
\end{equation}

The rotation matrix \({}_I^{C}R\in\text{SO(3)}\) is constructed from the basis \({}_I^{C}R=[\bm{c_x}\; \bm{c_y}\; \bm{c_z}]\), and the error rotation matrix is defined as \({}_B^{C}R\in\text{SO(3)}\). The attitude error vector can be obtained using the logarithmic map \(\ln (.): \text{SO(3)}\rightarrow\mathfrak{so}(3)\), and the vee operator \(.^\vee:\mathfrak{so}(3)\rightarrow \mathbb{R}^3 \):
\begin{equation}
\label{eq_rotation_vector}
    \begin{aligned}
        \eta=&\arccos{(\frac{\text{Tr}({}_B^CR)-1}{2})} \\
        \bm{e_\eta}=&\ln({}_B^CR)^\vee=\frac{\eta}{2sin(\eta)} ({}_B^CR-{}_C^BR)^\vee
    \end{aligned}
\end{equation}

The singularities due to small rotation angles or large rotation angles close to \(\pm\pi\) are handled properly. The attitude controller is then defined as:
\begin{equation}
    \label{eq_att_ctrl}
    \bm{u_\eta}=K_\eta\bm{e_\eta}+K_\omega\bm{e_\omega}
\end{equation}
with \(K_\eta\) and \(K_\omega\) being constant diagonal matrices.

Finally, we can calculate the collective thrust command $u_T$ by projecting the commanded acceleration vector to \(\bm{c_z}\):
\begin{equation}
\label{eq:uz_feedback_linearization}
\begin{aligned}
    u_T =& k_b {}^I\bm{a_d} \cdot \bm{c_z} \\
    k_b =&\frac{\hat{m}}{\mu_n\hat{k}_F}
\end{aligned}
\end{equation}
where the constants \(\hat{m}\), and \(\hat{k}_F\) are the estimated UAV mass, and the thrust coefficient from Eq. \eqref{eq_act_dynamics_time}, respectively. Note that in Eqs. \eqref{eq_outer_loop_ctrl} and \eqref{eq_rotation_vector} rotation matrices based on attitude and heading measurements are used, which have their own delay. Such delays are dropped for simplicity.

\subsection{Decoupled and Linearized Model}
Analysis of decoupled dynamics can be achieved by projecting the 3D space into a 2D space. Specifically for decoupling, we assume that \(\mathcal{F}_I:=\mathcal{F}_H\) and, without loss of generality, project on the plane defined by \(\bm{i_y}\cross \bm{i_z}\). The rotation around \(\bm{b_x}\) is indicated by the angle \(\theta\). All vectors and matrices used in this section are compatible with a 2D space, unless explicitly indicated otherwise. The resultant UAV dynamics along \(\bm{i_y}\) and \(\bm{i_z}\) becomes:
\begin{equation}
    \begin{aligned}
        {}^I\dot{p}_y=&{}^Iv_y\\
        {}^I\dot{v}_y=&\sin \theta \frac{F}{m} - (d_y\cos^2\theta+d_z\sin^2\theta){}^Iv_y \\
        &- (d_y\sin\theta\cos\theta+d_z\sin\theta \cos\theta){}^Iv_z\\
        {}^I\dot{p}_z=&{}^Iv_z\\
         {}^I\dot{v}_z=&\cos \theta \frac{F}{m}  - (d_y\sin\theta\cos\theta+d_z\sin\theta \cos\theta){}^Iv_y \\
         &- (d_y\sin^2\theta+d_z\cos^2\theta){}^Iv_z\\
         \dot{\theta}=&\omega\\
         \dot{\omega}=&\frac{1}{J_x}M_x-A_x(\cos\theta {}^Iv_y+\sin\theta) {}^Iv_z-B_x\omega
    \end{aligned}
\end{equation}

The components of \(\bm{c_z}\) vector from Eq. \eqref{eq_fb_lin_r_bc} become \(c_{z,x}=0\), \(c_{z,y}=\frac{{}^Ia_{d,y}}{a_d}=\sin\theta_d\), and \(c_{z,z}=\frac{{}^Ia_{d,z}}{a_d}=\cos\theta_d\) where \(a_d=\|{}^I\bm{a_d}\|\). Using \(\ln{{}_I^CR}\), the desired angle of the inner attitude loop becomes \(\theta_d=\atan\frac{{}^Ia_{d,y}}{{}^Ia_{d,z}}\). The same relation holds for thrust force generated so that \(\sin\theta=\frac{a_{F,y}}{a_F}\) and \(\cos\theta=\frac{a_{F,z}}{a_F}\) with \({}^I\bm{a_F}=\frac{F}{m}\bm{b_z}=[{}^Ia_{F,y} {}^Ia_{F,z}]^T\) and \(a_F=\|{}^I\bm{a_F}\|\). Hence, under feedback control given in the previous section, the angular closed loop dynamics, including actuator dynamics, become:
\begin{equation}
    \begin{aligned}
    \dot{\theta}=&\omega\\
     \dot{\omega}=&\frac{1}{J_x}M_x-A_x(\cos\theta {}^Iv_y+\sin\theta) {}^Iv_z-B_x\omega\\
     \dot{M}_x=&\frac{-M_x+k_{M,b_x}u_{b_x}(t-\tau_p)}{T_p}\\
     u_{b_x}(t-\tau_p)=&K_{\eta,x}(\theta_c(t-\tau_p)-\theta(t-\tau_p-\tau_{\eta m}))\\
     &+K_{\omega,x}(\omega(t-\tau_p-\tau_{\omega m}))
    \end{aligned}
\end{equation}
where \(k_{M,b_x}=k_F l_y /\sqrt{2}\). With the drop of the velocity dependent drag term, \(A_x(\cos\theta {}^Iv_y+\sin\theta) {}^Iv_z\) which is due to blade flapping, the angular dynamics become linear. The transfer function \(G_\theta(s)=\Theta(s)/\Theta_r(s)\) of the angular dynamics is given by:
\begin{equation}
\label{eq_attitude_tf}
    G_\theta(s)=\frac{K_\theta(K_{\eta,x}+sK_{\omega,x})e^{-\tau_\theta s}}{s(T_ps+1)(\frac{J_x}{B_x}s+1)+K_\theta(K_{\eta,x}+sK_{\omega,x})e^{-\tau_\theta s}}
\end{equation}
where \(K_\theta=\frac{k_{M,b_x}}{B_x}\), and \(\tau_\theta=\tau_p+\tau_{\eta m}\). The altitude loop dynamics are given by:
\begin{equation}
    \begin{aligned}
    {}^I\dot{p}_z=&{}^Iv_z\\
    {}^I\dot{v}_z=&\cos\theta a_F - (d_y\sin\theta\cos\theta+d_z\sin\theta \cos\theta){}^Iv_y \\
    &- (d_y\sin^2\theta+d_z\cos^2\theta){}^Iv_z\\
    \dot{a}_F=&\frac{\mu_n\frac{k_F}{m}u_T(t-\tau_p)}{T_p}
    \end{aligned}
\end{equation}

By substituting \(\cos\theta\) with
\begin{equation*}
    \cos(\mathcal{L}^{-1}\{G_\theta(s)\}\atan(\frac{{}^Ia_{d,y}}{{}^Ia_{d,z}}))
\end{equation*}
and approximating \({}_I^BR\approx I_{2\times 2}\) for the drag dynamics, the altitude closed loop dynamics become:
\begin{equation}
\label{eq_altitude_dyn_full}
    \begin{aligned}
    {}^I\dot{p}_z=&{}^Iv_z\\
    {}^I\dot{v}_z=&\cos(\mathcal{L}^{-1}\{G_\theta(s)\}\atan(\frac{{}^Ia_{d,y}}{{}^Ia_{d,z}})) a_F - d_z{}^Iv_z-g\\
    \dot{a}_F=&\frac{\frac{k_F}{m}\frac{\hat{m}}{\hat{k}_F}\|{}^I\bm{a_d}(t-\tau_p)\|}{T_p}-\frac{a_F}{T_p}
    \end{aligned}
\end{equation}

The above system is a multi-input-single-output (MISO) system with the inputs being \({}^I\bm{p}_{r}\), \({}^I\bm{v}_{r}\), and \({}^I\bm{a}_{r}\). The altitude loop becomes decoupled from the lateral motion loop during hover, i.e. when \({}^Ia_{d,y}=0\), at which the system becomes linear with the following transfer function (assumig perfect estimates provided by \(\hat{m}\), \(\hat{k}_F\), and \(\hat{a}_g\)):
\begin{equation}
\label{eq_altitude_tf}
    \frac{P_z(s)}{P_{z,r}(s)}=\frac{\frac{1}{d_z}(K_{p,z}+sK_{v,z}+s^2)e^{-\tau_ps}}{s(T_ps+1)(\frac{s}{d_z}+1)+\frac{1}{d_z}(K_{p,z}+sK_{v,z})e^{-\tau_zs}}
\end{equation}
where \(\tau_z=\tau_p+\tau_{pm}\). Similar to the altitude dynamics given in Eq. \eqref{eq_altitude_dyn_full}, lateral motion dynamics are given by:
\begin{equation}
\label{eq_lateral_decoup_nonlinear}
    \begin{aligned}
    {}^I\dot{p}_y=&{}^Iv_y\\
    {}^I\dot{v}_y=&\sin(\mathcal{L}^{-1}\{G_\theta(s)\}\atan(\frac{{}^Ia_{d,y}}{{}^Ia_{d,z}})) a_F - d_y{}^Iv_y\\
    \dot{a}_F=&\frac{\frac{k_F}{m}\frac{\hat{m}}{\hat{k}_F}\|{}^I\bm{a_d}(t-\tau_p)\|}{T_p}-\frac{a_F}{T_p}
    \end{aligned}
\end{equation}

But decoupling the above equation for the hovering case when \({}^Ia_{d,z}=\hat{a}_g\) is not possible without some assumptions, as changing the input \( {}^Ia_{d,y}\) changes both \(\sin\theta\) and \(a_F\). We can disregard the contribution of the \(a_F\) term due to two reasons:
\begin{enumerate}
    \item First, when \(\theta\approx0\) \(\left.\frac{\partial {}^I\dot{v}_y}{\partial \theta}\right\vert_{\theta=0}=\cos\theta a_F\), and \(\left.\frac{\partial {}^I\dot{v}_y}{\partial a_F}\right\vert_{\theta=0}=0\).
    \item Second, \(\sin\theta\) dynamics are faster than \(a_F\) dynamics. In fact both share the same propulsion dynamic and hence the angular dynamics are slower by \(\frac{k_{M,b_x}/B_x}{s((J_x/B_x)s+1)}\). The slower dynamics dominate the overall response.
\end{enumerate}

Hence it is possible to obtain the linearized lateral motion dynamics as:
\begin{equation}
\label{eq_lateral_tf}
    \frac{P_y}{P_{r,y}}=\frac{\frac{1}{d_y}\hat{a}_gN_\theta(s)(K_{p,y}+K_{d,y}s+s^2)}{s(\frac{s}{d_y}+1)D_\theta(s)+\frac{1}{d_y}\hat{a}_gN_\theta(s) (K_{p,y}+sK_{d,y})e^{-\tau_{pm}s}}
\end{equation}
where \(G_\theta(s)=\frac{N_\theta(s)}{D_\theta(s)}\). Obviously, the previous analysis can be directly applied to the other decoupled loops. In total, we get six decoupled SISO control loops: altitude, roll, pitch, yaw, lateral along \(\bm{i_x}\) and lateral along \(\bm{i_y}\).
\begin{remark}
The time delay in the numerator of Eqs. \eqref{eq_altitude_tf} and \eqref{eq_lateral_tf} corresponds to the position measurement delay present in Eq. \eqref{eq_outer_loop_ctrl}. The generated position, velocity, and acceleration references could be advanced by \(\tau_{pm}\) such that the time delay effect is cancelled, and a better tracking could be achieved.
\end{remark}

\begin{remark}
For the feedback linearization term in Eq. \eqref{eq_lateral_decoup_nonlinear} to be valid, the condition \(G_\theta(s)\approx1 \; \forall \omega_{\theta c} \geq \omega_{td}\) needs to be satisfied; where \(\omega_{\theta c}\) is the \(G_\theta(s)\) system bandwidth and \(\omega_{td}\) is the reference trajectory bandwidth.
\end{remark}

\begin{figure} [t]
\centering
  \includegraphics[width = \linewidth]{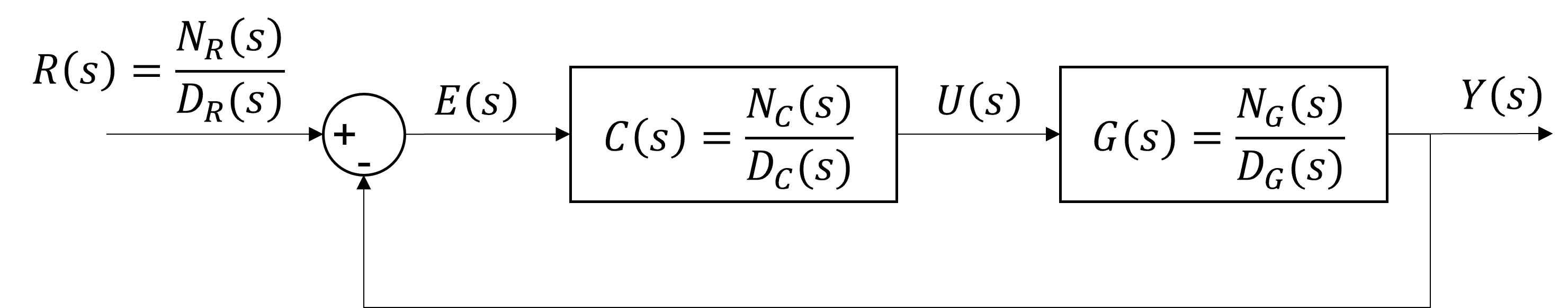}
\caption{A feedback control loop with unity dynamics in the feedback path. The dynamics of the error \(E(s)\) is made up of the dynamics of the system \(G(s)\), the controller \(C(s)\) and the reference signal \(R(s)\).}
\label{fig:fb}
\end{figure}
\section{Evaluation of Models Suitability Using Analytical Cost Functionals}\label{sec_eval_suitability_cost_func}
Most models used in the literature of UAVs are delay-free; where some researchers neglect actuator dynamics resulting in a second order system \cite{mellinger2011minimum,mahony2012multirotor,lee2013nonlinear,mohta2018fast,Jia2022wind}, where others account for actuator lag dynamics resulting in a system with relative degree three \cite{faessler2017differential,pounds2010modelling}. In what follows, we show that delay-free models are unsuitable for controller design due to the existence of a trivial solution that minimizes cost functional. In contrast, we will show that a trivial solution does not exist with time delay models.

\subsection{Analysis of Suitability of Low Order Models for Tuning}
For a SISO linear system of the form shown in Fig. \ref{fig:fb}, the closed loop error dynamics are given by:
\begin{equation}\label{eq_err_dyn}
    E(s)=\frac{D_C(s)D_G(s)N_R(s)}{D_R(s)(D_C(s)D_G(s)+N_C(s)N_G(s))}
\end{equation}
For the delay-free case, this can be rewritten as:
\begin{equation}
    E(s)=\frac{B(s)}{A(s)}=\frac{b_1s^{n-1}+b_2s^{n-2}+...+b_n}{a_0s^n+a_1s^{n-1}+...+a_{n}},\;a_0\neq0
\end{equation}
A controller design is often performed through minimization of the integral of the square error (ISE) performance index. It was shown in \cite{newton1957analytical} (we follow the notion used in \cite{marshall1992time}) that the ISE performance index of delay-free linear system can be found using Parseval's theorem as follows \cite{newton1957analytical}:
\begin{equation}
    J=\int_0^\infty{e^2(t) dt}= \frac{1}{2\pi i}\int_{-i\infty}^{+i\infty}{E(s)E(-s)}ds
\end{equation}
Exploiting symmetry, the integrand \(E(s)E(-s)\) can be written in the form \cite{marshall1992time}:
\begin{equation}
    E(s)E(-s)=\frac{B(s)B(-s)}{A(s)A(-s)}=\frac{F(-s)}{A(s)}+\frac{F(s)}{A(-s)}
\end{equation}
By equating polynomial coefficients we can find the following set of equations:
\begin{gather}
\begin{bmatrix}
a_1 & a_0 & 0 & 0 & . & . & . \\
a_3 & a_2 & a_1 & a_0 & 0 & . & . \\
a_5 & a_4 & a_3 & . & . & . & . \\
. &   &   &   &   &   &   \\
. &   &   &   &   &   &   \\
. &   &   &   &   &   &   \\
 &   &   &   & a_n & a_{n-1} & a_{n-2} \\
 &   &   &   &   &   & a_n
\end{bmatrix}
\begin{bmatrix}
f_1\\
f_2\\
.\\
.\\
.\\
\\
\\
f_n
\end{bmatrix}
=
\begin{bmatrix}
g_1\\
g_2\\
.\\
.\\
.\\
\\
\\
g_n
\end{bmatrix}
\end{gather}
where
\begin{equation}
    g_r=\frac{1}{2}\sum_{m=1}^{2r-1}{(-1)^{m-1}b_{2r-m}b_m}
\end{equation}
It was shown in \cite{newton1957analytical} that the cost functional evaluates to:
\begin{equation}
    J=\frac{f_1}{a_0}
\end{equation}

One may refer to \cite{marshall1992time} for a method on finding expressions for other symmetric and non-symmetric cost functionals of the form:
\begin{equation}
    \int_0^\infty{t^m(\frac{d^{(n)}e}{dt^{(n)}})^2}dt
\end{equation}
with \(m,n\in\mathbb{N}\). Most works in the literature ignore actuator dynamics, for which the same analysis in the sequel applies. Let us consider a third order model from the literature to describe attitude and altitude dynamics of UAVs with first order actuator dynamics:
\begin{equation}
\label{eq_low_order_model}
    G_l(s)=\frac{K_p}{s(T_ps+1)(T_qs+1)}
\end{equation}
with a PD controller:
\begin{equation}
    C_{pd}(s)=K_c+K_ds
\end{equation}
Then the step response of the closed loop error dynamics are defined by substituting Eq. \eqref{eq_low_order_model} in Eq. \eqref{eq_err_dyn} as:
\begin{equation}
\label{eq_error_dyn_simple}
    E_{f}(s)=\frac{(T_ps+1)(T_qs+1)}{T_ps^2+(K_dK_p+1)s+K_cK_p}
\end{equation}
for which, the ISE cost functional evaluates to:
\begin{equation} 
J =\frac{\frac{1}{2}T_pT_q(K_pK_d+1)+\frac{T_p+T_q}{2K_pK_c}-T_pT_q-\frac{(T_p+T_q)^2}{2}}{(T_p+T_q)(K_pK_d+1)-T_pT_qK_pK_c}
\end{equation}

Note that \(J\) is to be positive, otherwise the system is unstable. In fact, the positive requirement of the denominator of \(J\) is equivalent to the Routh-Hurwitz stability criteria \((T_p+T_q)(K_pK_d+1)>T_pT_qK_pK_c\) \cite{walton1984evaluation}. It follows that for the task of minimizing \(J\) with decision variables \(K_c\) and \(K_d\), a trivial solution exists:
\begin{equation}
\label{eq_trivial_cost}
    \min_{K_c,K_d} J=\lim_{K_c\to\infty,\,K_d\to\infty}J=\frac{T_pT_q}{2(T_p+T_q)},\;\;\text{s.t.}\, J>0
\end{equation}

Therefore, an attempt to minimize \(J\) by finding an optimal combination of \(K_c\) and \(K_d\) would lead to the noted trivial solution. For the case when the first order dynamics used to account for actuator dynamics are neglected, a trivial solution also exists with the optimal value of \(J\) being zero. The analysis given serves as evidence of unsuitability of the low order models similar to the one given in Eq. \eqref{eq_low_order_model} for PD controller design.

\subsection{Finding Analytical Cost Functional for Time Delay systems}
One cannot directly evaluate the integrand \(E(s)E(-s)\) using Parseval's theorem in the delay case due to the infinite number of poles in the left and right half-planes. Yet, it is possible to analytically evaluate a cost functional associated with linear systems with a single time delay through the method suggested in \cite{walton1986}. The method requires rewriting the closed loop error dynamics in the form:
\begin{equation}
\label{eq_closed_loop_delay_system}
    E(s)=\frac{B(s)+D(s)e^{-\tau s}}{A(s)+C(s)e^{-\tau s}}
\end{equation}
Let us use Cauchy's residue theorem to evaluate such integrals \cite{walton1986}:
\begin{equation}
\label{eq_poles_residue}
\begin{aligned}
    J=-\sum_k \text{res}_{s=s_k}E(s)
    (\frac{B(-s)A(s)-D(-s)C(s)}{A(-s)A(s)-C(-s)C(s)})
\end{aligned}
\end{equation}
for which the residues \(s_k\) are the roots of the equation:
\begin{equation}\label{eq_roots}
    L(s)=A(-s)A(s)-C(-s)C(s)=0
\end{equation}
The idea of the method is that the integrand is rearranged so that only a finite number of relevant poles are encircled in the integration contour \cite{walton1986}.

Eq. \eqref{eq_roots} assumes that the coefficient \(l_0\) associated with the term of the greatest power is non unity. For non-unity \(l_0\), Eq. \eqref{eq_poles_residue} has to be modified as follows:
\begin{equation}
\label{eq_poles_residue_mod}
\begin{aligned}
    J=-\frac{1}{l_0}\sum_k \text{res}_{s=s_k}E(s)
    (\frac{B(-s)A(s)-D(-s)C(s)}{A(-s)A(s)-C(-s)C(s)})
\end{aligned}
\end{equation}

Given a system of second order with integrator plus time delay (SOIPTD) dynamics:
\begin{equation}
    G(s)=\frac{K_pe^{-\tau s}}{s(T_ps+1)(T_qs+1)}
\end{equation}
with a PD controller, the error dynamics are written as:
\begin{equation}
\label{eq_error_dyn_toptd_pd_step}
    E(s)=\frac{s(T_ps+1)(T_qs+1)}{s(T_ps+1)(T_qs+1)+(K_pK_c+K_pK_ds)e^{-\tau s}} R(s)
\end{equation}

The matrices \(A(s)\), \(B(s)\), \(C(s)\), and \(D(s)\) of the SOIPTD system for a step reference signal become:
\begin{align}
    A(s)&=s(T_ps+1)(T_qs+1) \\
    B(s)&=(T_ps+1)(T_qs+1) \\
    C(s)&=K_pK_c+K_pK_ds \\
    D(s)&=0 \\
\end{align}
where we need to find the roots as in Eq. \eqref{eq_roots}:
\begin{equation}
    L(s)=-T_p^2T_q^2s^6 +(T_p^2+T_q^2)s^4+ (K_d^2K_p^2-1)s^2 -K_c^2K_p^2=0
\end{equation}
which is of sixth order. Because only even powers exist in the polynomial, we would substitute \(r=s^2\), solve a cubic equation, and then re-substitute \(s_{2i-1,2i}=\pm \sqrt{r_i}\). The mathematical expression for the roots is lengthy, so we used a symbolic solver to evaluate Eq. \eqref{eq_poles_residue}. The resultant Eq. \eqref{eq_poles_residue_mod} was analytic at all roots, i.e. it has simple roots, and hence the expression in Eq. \eqref{eq_poles_residue_mod} can be evaluated. For example, the first two summation terms in Eq. \eqref{eq_poles_residue_mod} corresponding to the residue evaluation for the poles \(s_1\) and \(s_2\) are given by:
\begin{strip}
\begin{equation*}%\resizebox{1.0\hsize}{!}{\(
    J_1=\frac{(T_p C_1 - 1)^2  (T_p C_1 + 1) (T_q C_1 - 1)^2  (T_q C_1 + 1)}{2 T_p^2  T_q^2  (C_2 + C_1) (C_3 + C_1) (C_2 - C_1) (C_3 - C_1) (e^{\tau C_1} (K_c K_p - K_d K_p C_1) - (T_p C_1 - 1) (T_q C_1 - 1) C_1)}
\end{equation*}
\begin{equation*}
     J_2=\frac{(T_p C_1 - 1)  (T_p C_1 + 1)^2 (T_q C_1 - 1)  (T_q C_1 + 1)^2}{2 T_p^2  T_q^2  (C_2 + C_1) (C_3 + C_1) (C_2 - C_1) (C_3 - C_1) (e^{-\tau C_1} (K_c K_p + K_d K_p C_1) + (T_p C_1 + 1) (T_q C_1 + 1) C_1)}
\end{equation*}
\end{strip}
where \(C_1\), \(C_2\), and \(C_3\) are constant terms independent of the delay value \(\tau\). Note the similar patterns that exist in \(J_1\) and \(J_2\) due to the fact that both stemmed from the root \(r_1\). Similarly, expressions for the other summation terms of  Eq. \eqref{eq_poles_residue_mod} can be found, but we omit them to converse space. Unlike the delay-free case, a trivial solution does not exist for the time delay cost functional.
\begin{comment}

But a prior condition to the existence and validity of the cost functional expression in Eq. \eqref{eq_poles_residue_mod} is the stability of the closed loop delay system. The stability of a retarded commensurate delay system, as for those of interest to this work given in Eqs. \eqref{eq_attitude_tf} and \eqref{eq_altitude_tf}, might be delay independent which needs to be investigated first. From \cite{gu2003stability} we use the following theorem:
\begin{theorem}
A system of the form
\begin{equation*}
    \dot{\bm{x}}(t)=A_0\bm{x}(t)+A_1\bm{x(t-\tau)}
\end{equation*}
is stable independent of delay if and only if (i) \(A_0\) is stable, (ii) \(A_0+A_1\) is stable, and (iii) \(\rho ((j\omega I - A_0)^{-1}A_1)<1\), \(\forall \omega > 0\). Here \(\rho(.)\) denotes the spectral radius of a matrix.
\end{theorem}

With the proof provided in \cite{gu2003stability}. The delay-free matrix \(A_0\) of the systems defined by Eqs. \eqref{eq_attitude_tf} and \eqref{eq_altitude_tf} has the form:
\begin{equation}
    A_0=\begin{bmatrix}
0 & 1 & 0\\
0 & 0 & 1\\
0 & a_{2} & a_{3}
\end{bmatrix}
\end{equation}
which is unstable, and hence the stability of the closed-loop systems \eqref{eq_attitude_tf} and \eqref{eq_altitude_tf} is delay dependent.
\end{comment}

But a prior condition to the existence and validity of the cost functional expression in Eq. \eqref{eq_poles_residue_mod} is the stability of the closed loop delay system. The conditions on asymptotic stability of the time delay system given in Eq. \eqref{eq_closed_loop_delay_system} can be analyzed by considering the characteristic equation given by:
\begin{equation}
\label{eq_ce_delay}
    F(s,h)\equiv A(s)+C(s)e^{-\tau s}=0
\end{equation}
for which the number of roots is infinite due to the delay term. If all of these roots are in the left half of the complex s-plane, then the system is stable, otherwise it is not. We can check the location of the roots by first checking the delay-free system:
\begin{equation}
    F(s,0)\equiv A(s)+C(s)=0
\end{equation}
which we already checked for the system in Eq. \eqref{eq_error_dyn_simple} and found it stable for \((T_p+T_q)(K_pK_d+1)>T_pT_qK_pK_c\). The delay-free stable system could become unstable if the introduction of the time delay causes some poles to cross the left half s-plane to the right half at \(s=\pm j\omega_c\). Substituting these crossing points into Eq. \eqref{eq_ce_delay} gives the two equations:
\begin{equation}
    \begin{aligned}
        A(j\omega_c)+C(j\omega_c)e^{-j\omega_c}&=0\\
        A(-j\omega_c)+C(-j\omega_c)e^{j\omega_c}&=0
    \end{aligned}
\end{equation}
from which the exponential terms can be reduced to yield:
\begin{equation}
    W(\omega_c^2)\equiv A(j\omega_c)A(-j\omega_c)-C(j\omega_c)C(-j\omega_c)=0
\end{equation}
The existence of a solution for \(\omega_c^2\) in  the above equation corresponds to the existence of a delay value \(\tau_c\) which would lead to instability. Substituting \(q_c=\omega_c^2\) we find the cubic polynomial corresponding to the system in Eq. \eqref{eq_error_dyn_toptd_pd_step}:
\begin{equation}
\begin{aligned}
       W(q_c)\equiv T_pT_qq_c^3+&(T_p^2+4T_pT_q+T_q^2)q_c^2+(1-K_p^2K_d^2)q\\
       &-K_p^2K_c^2=0 
\end{aligned}
\end{equation}
Positive real polynomial roots exist if sign changes occur in any of the first column entries of the Routh-Hurwitz table. The first column entries are (in order): \(T_pT_q\), \(T_p^2+4T_pT_q+T_q^2\), \(1-K_p^2K_d^2+\frac{K_p^2K_c^2T_pT_q}{T_p^2+4T_pT_q+T_q^2}\), and \(-K_p^2K_c^2\) which indicates that a real root exists regardless of the model and controller parameters. This leads to the conclusion that the introduction of delay would lead to instability, for a finite value of \(\omega_c\). This is fundamentally different from the qualitative behavior of the delay-free system in Eq. \eqref{eq_error_dyn_simple}. The value of delay \(\tau_c\) that corresponds to the crossing frequency \(\omega_c\) is given by:
\begin{equation}
    \begin{aligned}
    \sin{\omega_c\tau_c}&=\Im{\frac{A(j\omega_c)}{C(j\omega_c)}}\\
    &=\frac{(\omega_c-T_pT_q\omega_c^3)K_pK_c+K_pK_d(T_p+T_q)\omega_c^3}{K_p^2K_c^2+K_p^2K_d^2\omega_c^2}\\
    \cos{\omega_c\tau_c}&=\Re{\frac{-A(j\omega_c)}{C(\omega_c)}}\\
    &=\frac{(T_p+T_q)\omega_c^2K_pK_c+(T_pT_q\omega_c^3-\omega)K_pK_d\omega_c}{K_p^2K_c^2+K_p^2K_d^2\omega_c^2}
    \end{aligned}
\end{equation}
which yields infinite solutions of \(\tau_c\). But we are interested in the smallest solution \(\tau_0\) corresponding to the first crossing of poles across the imaginary axis as this is guaranteed to be a destabilizing crossing, i.e. poles cross to the right-hand side of the s-plane, since the delay-free system is stable. Obtaining an analytical expression for \(\tau_0\) is possible through the Nyquist stability criterion, where the first crossing of the poles to the right-hand side of the s-plane corresponds to:
\begin{equation}
    \angle \tau_0=\angle\frac{1}{\omega_c}(\atan\frac{K_d\omega_c}{K_c}-\atan T_p\omega_c-\atan T_q\omega_c+\frac{\pi}{2})
\end{equation}

Note that for the delay-free case with \(\tau_0=0\) the above equation can be satisfied by the trivial setting of \(K_d\rightarrow \infty\) which leads to \(\omega_c\rightarrow \infty\) such that the system is always stable, regardless of any choice of finite model parameters. This conclusion is in agreement with the one found in Eq. \eqref{eq_trivial_cost} where the existence of solutions to the controller parameters such that they tend to infinity hinders the delay-free models unusable in practice.

\section{Identification of Model Parameters Through DNN-MRFT} \label{sec:dnnmrft}
The DNN-MRFT approach first proposed in \cite{Ayyad2020} for identification and near optimal controller tuning for linear processes has shown very promising results for UAV control. Given the underactuated and nonlinear behavior of overall UAV system, controller design and tuning is quite complex and challenging task, especially considering that the system is inherently unstable. DNN-MRFT approach is used to excite stable oscillations for each loop in the control system to record system behavior experimentally and subsequently a pre-trained DNN is used for identification. 

\begin{table*}[t!]
\begin{center}
\caption{Range of model parameters, optimal value of MRFT parameter \(\beta^*\), and the test set reported maximum and average values of the relative sensitivity function shown in Eq. \eqref{eq:rel_sens_function} for every linearized dynamic model used. Note that \(\beta^*\) value for the underactuated lateral loops depends on attitude identification, and it is usually close to zero.}
\label{tab:dnn_mrft_parameters}
\bgroup
\def\arraystretch{1.5}
\begin{tabular}{ |l|l|l|l|l| }
 \hline
  Dynamics & Range of model parameters & \(\beta^*\) & Average \(J\) & Maximum \(J\)  \\
 \hline
 Attitude & \(T_{prop_{b_x}},T_{prop_{b_y}}\in [0.015,0.3];\;T_{\gamma_x},T_{\gamma_y}\in [0.2,2];\;\tau_{b_x},\tau_{b_y}\in [0.0005,0.15]\) & -0.73 & 0.53\% & 3.51\% \\
 \hline
 Altitude & \(T_{prop_z}\in [0.015,0.3];\;T_{\lambda_z}\in [0.2,2];\;\tau_z\in [0.0005,0.15]\) & -0.71 & 0.3\% & 5.03\% \\
 \hline
  Lateral & \(T_{\lambda_x},T_{\lambda_y}\in [0.2,6];\;\tau_x,\tau_y\in [0.0005,0.15]\) & Varies & -0.19\% & 4.91\% \\
 \hline
\end{tabular}
\egroup
\label{tab:parameters_mapping}
\end{center}    
\end{table*}

\subsection{Problem Formulation}

The whole UAV system is divided into attitude, altitude and lateral control loops which are linearized and denoted by \(G(s, d_i)\), where $s$ is the Laplace variable and \(d_i \in D)\) denotes the unknown parameters associated with the system from the parameter space D. For brevity we drop the Laplace variable $s$ and use \(G(d_i)\) for all linear processes involved. The parameter space \(D= \tau \times T_{prop} \times T_{\lambda}\times K_{eq}\) consists of the time delay, time constants and equivalent gain of the system. The limits of these parameters for each control loop are defined in Table \ref{tab:parameters_mapping} forming the boundaries of $D$. Now this parameter space is discretized to get a finite number of set of parameters for each system and denoted by $\check{D}$. If the features' subspace is defined by \(S\subset \mathbb{R}^{n\times m}\), given that $n$ is the number of elements in a feature vector and $m$ is the number of feature vectors in each observation then \(s_i \in S\) is single observation from experimental data. Then the goal for the DNN is to learn the inverse map \(s_i\mapsto d_i=M^{-1}(s_i)\), so that it can infer the unknown process parameter set $d_i$ from the experimental data $s_i$. Training is done using simulations of linear processes, hence for data generation some tunable parameters \(\zeta\) are also used to simulate noise and biases present in the actual system. Therefore the data generation function can be given by \((d_i,\zeta)\mapsto s_i=M(d_i,\zeta)\) for the particular simulated linear process \(G(d_i)\).  

\begin{comment}

Let \(\tau \in [\tau_{min},\tau_{max}]\), \(T_{prop} \in [T_{prop,min},T_{prop,max}]\), \(T_{\lambda} \in [T_{\lambda,min},T_{\lambda,max}]\), and \(K_{eq} \in [K_{eq,{min}},K_{eq,{max}}]\), with \(D= \tau \times T_{prop} \times T_{\lambda}\times K_{eq}\) denoting the domain of the unknown parameters characterizing a linear system \(G(s,d_i)\), (where $s$ is the Laplace variable, and will be dropped for brevity throughout the rest of this paper) where \(d_i \in D\). Let us define the features' subspace \(S\subset \mathbb{R}^{n\times m}\), where \(n\) denotes the number of elements in each feature vector, and \(m\) denotes the number of feature vectors. Then the map \((d_i,\zeta)\mapsto s_i=M(d_i,\zeta)\), where \(s_i \in S\) and \(\zeta\) denotes tunable data generation parameters, defines a data generation function that acts on the process \(G(d_i)\). Our goal now is to find an inverse map, \(s_i\mapsto d_i=M^{-1}(s_i)\), to infer the unknown process parameters from experimental data \(s_i\).

\end{comment}

\subsection{Data Generation Through NP-MRFT}
The data generation function \(M\) we use is a variant of the MRFT proposed in \cite{boiko2012nonparametricbook}. The challenge with the application of MRFT is that it suffers from false switchings due to low frequency noises present in vision based UAVs. Recently, the noise protected MRFT (NP-MRFT) was proposed in \cite{hay2021unified} to mitigate the effect of noise on the switching phase of MRFT. Specifically, NP-MRFT observes the error signal for a period of \(\tau_{obs}\) after every error signal peak and anti-peak to avoid switching at local peaks. The NP-MRFT is given by \cite{hay2021unified}:
\begin{comment}
\begin{multline}\label{eq_mrft_algorithm}
u_M(t)=\\
\left\{
\begin{array}[r]{l l}
h\; &:\; e(t) \geq b_1\; \lor\; (e(t) > -b_2 \;\land\; u_M(t-) = \;\;\, h)\\
-h\; &:\; e(t) \leq -b_2 \;\lor\; (e(t) < b_1 \;\land\; u_M(t-) = -h)
\end{array}
\right.
\end{multline}
\end{comment}

\begin{equation}
%\begin{split}
\label{eq_np_mrft_algorithm}
u^{(k)}_M(t)=\left\{
\begin{split}
h \;\;\;\; \text{if} \!\begin{multlined}[t]  (e(t) \geq b^{(k)}_1 \land t>t^{(k)}_{gmin}+\tau_{obs})\\
\lor\; (e(t) > -b^{(k)}_2 \;\land\; u^{(k)}_M(t-) = \;\;\, h )
\end{multlined}\\
-h \;\; \text{if} \!\begin{multlined}[t] (e(t) \leq -b^{(k)}_2 \land t>t^{(k)}_{gmax}+\tau_{obs}) \;\\
\lor\; (e(t) < b^{(k)}_1 \;\land\; u^{(k)}_M(t-) = -h )
\end{multlined}
\end{split}
\right.
%\end{split}
\end{equation}
where \(k\) denotes the \(k\)-th period of the NP-MRFT induced oscillations, $b_1$ and $b_2$ are defined by the equations:
% The following 2 equations are added on Jan. 8, 2022
\begin{equation}
\label{eq_modif_sw2a}
b^{(k)}_2=\beta e^{(k)}_{gmax}+a_n(1-\beta)
\end{equation}
\begin{equation}
\label{eq_modif_sw1a}
b^{(k)}_1=-\beta e^{(k)}_{gmin}+a_n(1-\beta)
\end{equation}
where the global maximum \(e^{(k)}_{gmax}(t)\) and global minimum \(e^{(k)}_{gmin}(t)\) are defined by:
\begin{equation}
\label{eq_glob_max}
e^{(k)}_{gmax}(t)=
\left\{
\begin{array}[r]{l l}
e^{(k)}_{max}(t) \; \text{if} \; e^{(k)}_{max}(t)=e^{(k)}_{max}(t-\tau_{obs})\\
e^{(k-1)}_{gmax}(t) \; \text{otherwise}
\end{array}
\right.
\end{equation}
and
\begin{equation}
\label{eq_glob_min}
e^{(k)}_{gmin}(t)=
\left\{
\begin{array}[r]{l l}
e^{(k)}_{min}(t) \; \text{if} \; e^{(k)}_{min}(t)=e^{(k)}_{min}(t-\tau_{obs})\\
e^{(k-1)}_{gmin}(t) \; \text{otherwise}
\end{array}
\right.
\end{equation}

and lastly the local maximum and minimum errors are given by: 
\begin{equation}
\label{eq_cur_max}
e^{(k)}_{max}(t)=
\left\{
\begin{array}[r]{l l}
\max_{t \in [t^{(k)}_{0+},t]} e(t) \; \text{if} \; e(t)>0\\
0 \; \text{otherwise}
\end{array}
\right.
\end{equation}
and
\begin{equation}
\label{eq_cur_min}
e^{(k)}_{min}(t)=
\left\{
\begin{array}[r]{l l}
\min_{t \in [t^{(k)}_{0-},t]} e(t) \; \text{if} \; e(t)<0\\
0 \; \text{otherwise}
\end{array}
\right.
\end{equation}
where $t^{(k)}_{0+}:=\{ t \;|\; e(t^{(k)}_{0+})=0 \; and \; e(t)>0 \; \forall \; t>t^{(k)}_{0+}\}$ is the time of the last zero crossing by increasing $e(t)$, and $t^{(k)}_{0-}:=\{ t \;|\; e(t^{(k)}_{0-})=0 \; and \; e(t)<0 \; \forall \; t>t^{(k)}_{0-}\}$ is the time of the last zero crossing by decreasing $e(t)$. For the choice of $\tau_{obs}>T_n\equiv\frac{2 \pi}{\Omega_n}$ with \(\Omega_n\) being the noise frequency, the NP-MRFT mitigates false switchings and hence the response becomes equivalent to MRFT applied to noise free error signal. Due to this equivalency and for compatibality with the literature, we refer to NP-MRFT as MRFT even though NP-MRFT is what is implemented on-board.

The use of MRFT for data generation has multiple advantages over other methods. Firstly, it is a closed-loop method making the system stable and robust to external disturbances during the data generation phase (refer to \cite{ayyad2021tcst} for a proof of orbital stability of the generated oscillations). Secondly, it exploits the gain scale property of linear systems, thus eliminating the need for including \(K_{eq}\) in the unknown parameters' domain \(D\), which greatly reduces the numerical complexity of the problem. We use the error input to MRFT \(e(t)\) and the MRFT output \(u_M(t)\) as data vectors for training; hence \(S\subset\mathbb{R}^{n\times 2}\). The parameter \(n\) defines the length of the data vector, and it is selected to fit the largest steady-state oscillation period from all members of \(\check{D}\). A suitable value of the MRFT parameter \(\beta\in[-1,1]\) is found through the process of the design of optimal non-parametric tuning rules \cite{boiko2012nonparametricbook}, resulting in the optimal data generation parameter \(\beta^*\). For simplicity, we refer to \(M(d_i,\beta^*)\) by \(M(d_i)\).

A summary of the DNN-MRFT tunable parameters used in this work for attitude, altitude and lateral loops is given in Table \ref{tab:dnn_mrft_parameters}.

\subsection{Process Classification based on DNN}
\label{sec:dnn_class}
A DNN is used to learn the inverse map \(M^{-1}\) to identify the unknown system parameters $d_i$ from the experimental MRFT generated data $s_i$. This is fundamentally a regression problem, however it would require huge training dataset and training times, especially considering that we use multiple DNNs. Hence we use a discretized parameter space \(\check{D}\), where we discretize based on Equation \ref{eq_disc_parameters_domain} to ensure the performance loss does not exceed certain limit. Now, the objective of DNN is changed from regression to process classification by learning \(\check{M}^{-1} : \check{S}\mapsto \check{D}\) mapping. The discretization function is given below :

\begin{equation}\label{eq_disc_parameters_domain}
    \Delta(D,J^*)=\{\check{D} : J(\check{d}_i,\check{d}_j)<J^* \land \{\check{d}_i,\check{d}_j\}\in \check{D}\}
\end{equation}

where \(J(d_i,d_j)\) is relative sensitivity function from \cite{rohrer1965sensitivity}, defined as:

\begin{equation}\label{eq:rel_sens_function}
    J(d_i, d_j) = \frac{Q(C(d_j) G(d_i)) - Q(C(d_i) G(d_i))}{Q(C(d_i) G(d_i))} \cross 100\%
\end{equation}

Given that $C(d_i)$ is the optimal controller for process $G(d_i)$, we can say that $J(d_i, d_j)$ provides a measure for performance deterioration when we apply the optimal controller of process $G(d_j)$ to process $G(d_i)$ compared to the optimal controller $C(d_i)$ performance. Note that $Q(\cdot)$ gives the performance index such as integral square error (ISE) on step response of the given system. Now $J^*$ has to be chosen properly to avoid over or under-discretization. In our case, we have used $J^* = 10\%$ which provides a good balance.

\begin{figure}
    \centering
    \includegraphics[width=\linewidth]{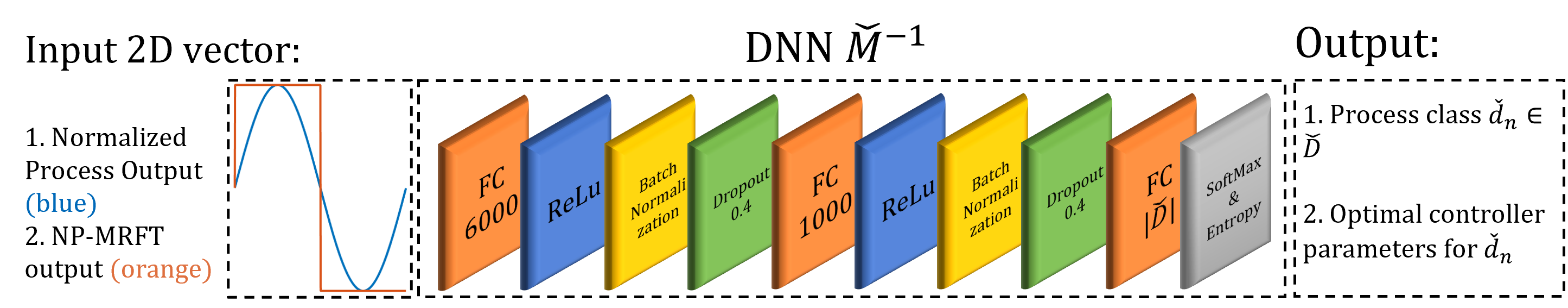}
    \caption{Details of the classification DNN architecture, its inputs, and its outputs. The inputs are the process output variable and the NP-MRFT controller action, while the output is the identified class \(\check{d}_n\). The last fully connected (FC) layer has a size of the number of classed (discretized processes in \(\check{D}\))}
    \label{fig:dnnarch}
\end{figure}

The DNN architecture has been summarized in Figure \ref{fig:dnnarch}. Multiple fully connected (FC) layers have been used with ReLu activation function. Batch Normalization is also used for better training while dropout layers with a rate of $40\%$ have been used to avoid over-fitting. Finally a modified Softmax function is used for classification given by:

\begin{equation} \label{eq_modified_softmax}
p_{i} = \frac{e^{\gamma_{iT}\cdot a_{i}}}{ \sum_{j=1}^{N}  e^{\gamma_{jT}\cdot a_{j}}}
\end{equation}

where $a_i$ is the output from final FC layer, \(\gamma_{iT}=1+J(d_i,d_T)\) and $p_i$ gives the classification probability. Since the cross-entropy loss function \(L=-\sum_{i=1}^{N} y_i \log{(p_i)}\) is used for training the DNN, the partial derivative with respect to $a_i$ used for backpropagation is given by:

\begin{equation} \label{eq_modified_softmax_backprop}
\frac{\partial L}{\partial a_{i}} = \gamma_{iT} \times (p_{i} - y_{i})
\end{equation} 

where $y_i = T$ is the ground truth class label. Such modification is necessary to penalize the misclassifications based on performance deterioration, rather than equally penalizing all. More details on training and analysis can be found in \cite{Ayyad2020}.

As we know that the under-actuated lateral loops dynamics are coupled with attitude dynamics, hence the identification and tuning of attitude loops directly affect the lateral loops which means that we have to perform DNN-MRFT based identification and tuining for attitude loops first and then proceed to lateral loops. Consequently, the mapping to be learned for lateral loops will be defined as \((\check{d}_{l,i},\check{d}_i)\mapsto \check{s}_{l,i}=\check{M}_l(\check{d}_{l,i},\check{d}_i)\) with \(\check{d}_{l,i}\in \check{D}_l\) for lateral dynamics parameters, and \(\check{d}_i\in \check{D}\) corresponding attitude dynamics parameters. This also entails that for each class of attitude loop, we train a separate DNN for the lateral loop. Although it might be possible to generalize the DNN for all attitude loops, however it is likely to result in bigger DNN which may not be suitable for real-time application on embedded processors.

\begin{figure}
    \centering
    \includegraphics[width=\linewidth]{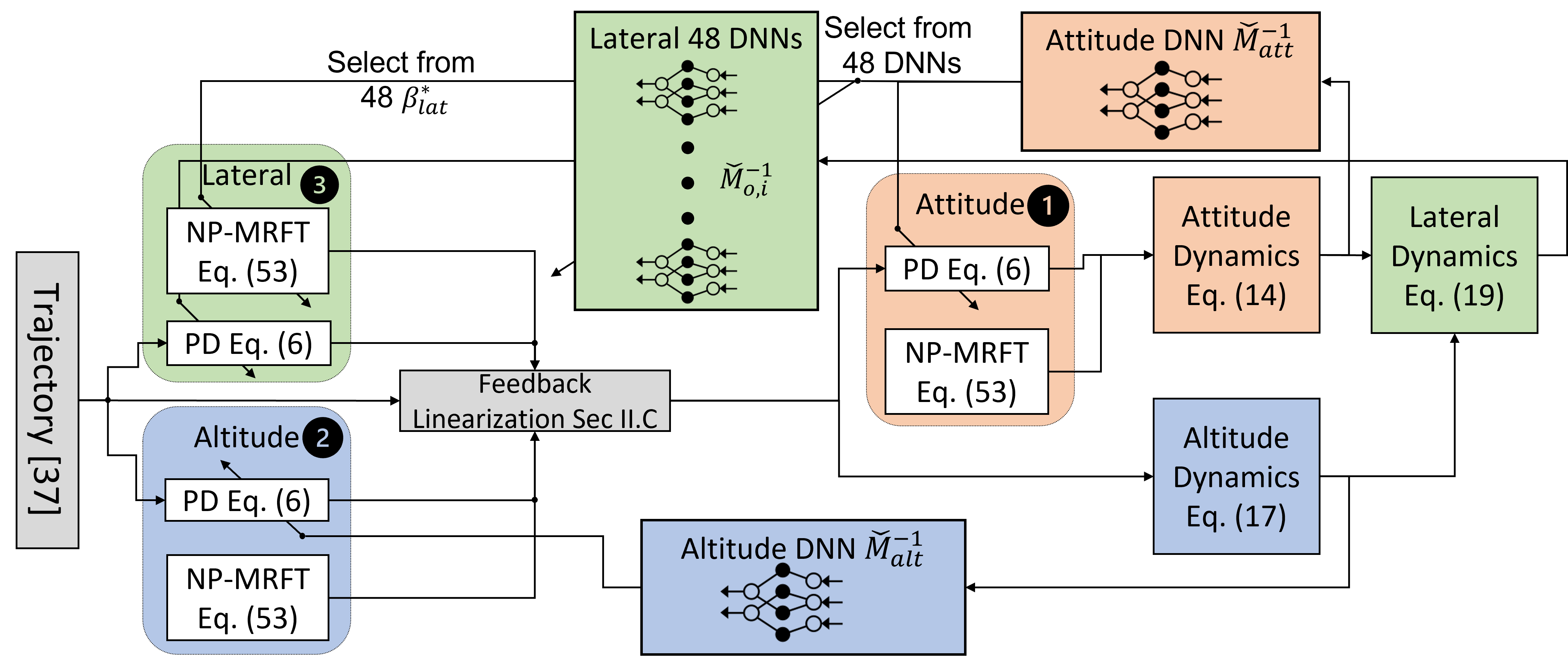}
    \caption{The arrangement of the DNNs and NP-MRFT used for identification. The identification sequence followed is shown. We switch for NP-MRFT during the identification phase, and back to the corresponding PD controllers in the normal operation phase. Note how the attitude DNN output selects the appropriate DNN and \(\beta^*\) for the lateral dynamics identification}
    \label{fig:dnnmrft_arrangement}
\end{figure}

\subsection{Controller Tuning}
According to the time delay model presented in Section \ref{sec_model_control} some delay due to sensors and digital circuits is present in the feedback path, while the other part of delay is in the forward path mainly due to digital processing and delay in the electronic actuation system. A limitation of DNN-MRFT is that it reveals total loop dynamics without differentiating between dynamics in the forward path and the feedback path. Yet this does not affect the process of optimal parameter tuning. This can be seen from the relation between the time delay cost functional evaluation when all delay is allocated in the forward path \(J_F\), and when all of it is allocated to the feedback path \(J_B\). For example, when a step function is used in the error function, the following holds:
\begin{equation}
    J_F=\tau+J_B
\end{equation}
as the time delay term \(\tau\) is independent of the controller parameters we get:
\begin{equation}
    \text{argmin}_{K_c,K_d} J_F\equiv\text{argmin}_{K_c,K_d}J_B
\end{equation}
and thus controller parameters tuning is independent of where this delay is allocated. Note that by the structure assumed in Fig. \ref{fig:fb} all the delay is in the forward path of the loop.

For each model characterized by \(\check{d}_i\), optimal controller parameters can be found through minimizing Eq. \eqref{eq_poles_residue_mod}. We found that the closed loop system with optimal gains to be too sensitive to parametric variations, which yields extremely dense \(\check{D}\). To reduce the relative sensitivity of the system we introduce gain margin and phase margin specifications, where these specifications can be represented by the following constraints:
\begin{align}
    \text{min}_{q_i}J(q_i,d_i), \;\; \text{s.t. } \gamma_m<\gamma_{ms} \land \phi_m<\phi_{ms} 
\end{align}
where \(q_i\) is the set of controller parameters, \(\gamma_m\) and \(\phi_m\) are gain and phase margins respectively, and \(\gamma_{m,s}\) and \(\phi_{m,s} \) are the gain and phase margin specifications respectively. Details on enforcing gain and phase margin specifications in the optimization process can be found in \cite{boiko2012nonparametricbook}. From our previous work in \cite{Ayyad2020} and \cite{ayyad2021tcst} we found \(\phi_{m,s}=20^{\circ}\) to be a suitable choice that provides adequate compromise between performance, robustness, and numerical complexity.

\subsection{Identification Sequence}
\label{sec:id_seq}
The identification and tuning of the decoupled control loops in Equations \eqref{eq_attitude_tf}, \eqref{eq_altitude_tf} and \eqref{eq_lateral_tf} must be performed in certain sequence. For example, the dynamics in Eq. \eqref{eq_lateral_tf} depends on the closed loop attitude dynamics in Eq. \eqref{eq_attitude_tf}. Therefore, the attitude dynamics need to identified before the lateral loop dynamics. On the other hand, the attitude and altitude dynamics in Eqs. \eqref{eq_attitude_tf} and \eqref{eq_altitude_tf} respectively, are independent of any other dynamics. But a less obvious dependency of lateral dynamics on altitude exists which can be examined in Eq. \eqref{eq_lateral_decoup_nonlinear}. Specifically, the term \(\|{}^I\bm{a_d}(t-\tau_p)\|\) expands to \(\|{}_H^IR(K_p{}_I^HR\bm{e_p}+K_v{}_I^HR\bm{{e}_v}+K_i{}_I^HR\bm{\bar{e}_p})+{}^I\bm{a_r}+\hat{a}_g\bm{i_z}\|\) which contains controller parameters of the altitude dynamics. Hence, it is necessary to perform identification and tuning on the altitude dynamics prior to performing identification on lateral dynamics. Figure \ref{fig:dnnmrft_arrangement} summarizes the DNN-MRFT architecture and how the feedback controllers, NP-MRFT, DNNs, and physical dynamics being integrated.

\section{Experimental Results}\label{sec_exp_results}

\begin{comment}
\begin{figure} [t]
\centering
  \includegraphics[width = \linewidth]{figures/changing_light.png}
\caption{The UAV performance under changing lighting conditions}
\label{fig:qdrone}
\end{figure}

\begin{figure} [t]
\centering
  \includegraphics[width = 0.7\linewidth]{figures/slalom.png}
\caption{High speed Slalom maneuver test}
\label{fig:qdrone}
\end{figure}
\end{comment}

\subsection{Experimental Setup}
The experimental setup consists of three main parts: the UAV, the ground control PC, and the motion capture system. The UAV we use is the DJI F550 hexarotor kit fitted with six DJI E600 propulsion systems, which accept PWM ESC commands at the rate of 200 Hz. The Raspberry Pi 3B+ with Navio2 is used as a flight controller, and the NVIDIA Jetson TX2 with Orbitty carrier board is used on-board the UAV to interface with ZED mini camera and run the visual odometry pipeline. The platform measures \(\text{79}\times\text{72}\times\text{27}\) cm, weighs 2.74 kg, and have the rotational inertias of \(I_x=0.031\text{ kg}\cdot \text{m}^\text{2}\), \(I_y=0.030\text{ kg}\cdot \text{m}^\text{2}\), and \(I_z=0.052\text{ kg}\cdot \text{m}^\text{2}\). We use positional and yaw estimates from visual odometry (VO) implementation provided by ZED SDK over ROS. Velocity estimates are obtained using a Kalman filter fusing acceleration from IMU and position from VO. Orientation (roll and pitch) and angular rates are estimated using IMU measurements. The XSens MTi670 is used as an inertial sensor, OptiTrack motion capture system with Prime13 cameras is used to provide ground truth measurements. A video showing the experiments can be viewed at https://youtu.be/XdLdL9QeSWE.

\subsection{Identification Results and Tuning}
For controller tuning, DNN-MRFT approach as described in Section \ref{sec:dnnmrft} has been implemented. Firstly, we excite MRFT oscillations separately for each of roll and pitch loops, assuming the two are decoupled and following the sequence described in Section \ref{sec:id_seq}. The MRFT output and the measured process variable signals are then passed through the trained DNN to obtain SOIPTD model parameters and optimal controller gains for each of the loops. Note that the tuning of these attitude controllers is solely based on the measurements from IMU, hence it is independent of the positional measurement source (i.e. motion capture or VO). Once the attitude controllers are tuned, we proceed with DNN-MRFT to find the altitude model parameters and the corresponding optimal controller parameters. The DNN-MRFT identification is then continued for each of the lateral loops along \(\bm{h}_x\) and \(\bm{h}_y\) to obtain optimal tuning. The obtained identification and control parameters are summarized in Table \ref{tab:id_cont}. The identified model parameters through DNN-MRFT do not necessarily correlate with the true physical parameters due to the map \(M\) being surjective. For example, it can be noticed from Table \ref{tab:id_cont} that \(T_p\) for roll and pitch is not the same despite using the same propulsion system in both loops. But what is important from a feedback control perspective is that both roll and pitch dynamics are similar, which is reflected by the similar tuning of \(K_c\) and \(K_d\) for both loops.

Note that the propulsion system gains \(k_F\) and \(k_M\) in Eq. \eqref{eq_act_dynamics_time} are dependent on the battery voltage, which directly affects the closed loop gain \(K_p\). \(K_p\) is at maximum when the batteries are in full charge, and its value drops as the battery voltage drops. To ensure stability over the whole battery voltage range, we performed system identification when the battery was fully charged.

\begin{table}[ht]
    \centering
    \resizebox{0.49\textwidth}{!}{
    \begin{tabular}{|c|c|c|c|c|c|c|c|}
        \hline
         Loop & $K_c$ & $K_d$ & $K_p$ & $T_p$ & $T_1$ & $T_2$ & $\tau$ \\
         \hline
         roll & 0.411 &	0.066 &	76.87 & 0.071 & 0.276 & - & 0.02 \\
         pitch & 0.434 & 0.0652 & 146.58 & 0.0492 & 0.496 & - &	0.0175 \\
         altitude & 12.19 & 5.56 & 1.838 & 0.135 & 1.682 & - & 0.06 \\
         lateral \(\bm{h}_x\) & 7.35 & 3.93 & 0.827 & - & - & 0.867 & 0.139 \\
         lateral \(\bm{h}_y\) & 8.02 & 4.29 & 0.758 & - & - & 0.867 & 0.139 \\
         \hline
    \end{tabular}
    }
    \caption{DNN-MRFT based system identification results. $K_c$ and $K_d$ are proportional and derivative controller gains, respectively. The process gain $K_p$ resembles \(K_\theta\) from Eq. \eqref{eq_attitude_tf} for the attitude loops, \(\frac{1}{d_z}\frac{k_F}{m}\frac{\hat{m}}{\hat{k}_F}\) from Eqs. \eqref{eq_altitude_dyn_full} and \eqref{eq_altitude_tf} for the altitude loop, and \(\frac{1}{d_y}\hat{a}_gK_\theta K_\eta\) from Eq. \eqref{eq_lateral_tf} for the lateral loops. The propulsion time constant \(T_p\) in the attitude and altitude loops is shown in Eqs. \eqref{eq_attitude_tf} and \eqref{eq_altitude_tf}. The time constant \(T_1\) associated with rotational drag represents \(\frac{J_x}{B_x}\) from Eq. \eqref{eq_attitude_tf}. Also, the time constant \(T_1\) is used to represent the translational drag along the altitude loop represented by \(\frac{1}{d_z}\) from Eq. \eqref{eq_altitude_tf}. The time constant $T_2$ is used to represent the lateral motion drag term \(\frac{1}{d_y}\) from Eq. \eqref{eq_lateral_tf}. Finally, $\tau$ represents the total loop delay.}
    \label{tab:id_cont}
\end{table}

\subsection{Validation of Model Adequacy}
\begin{comment}
The system identification approach as described in Section \ref{sec:dnnmrft} is mainly formulated to guarantee near optimal performance rather than identifying exact parameters of the system. Considering the parameter space that has been discretized 
based on least-worst deterioration criteria may often result in significantly different system parameters due to big discretization steps where the performance using certain controller doesn't deteriorate much with change in system parameters. Especially, for the lateral loop we found that the performance becomes less sensitive to the change in time constant hence we have a big discretization step.

In aforementioned cases, it is no longer possible to compare the simulation and experimental results for validation. Hence, instead we propose to validate the behavior of the system by comparing the closed loop behaviour of the controlled system. We calculate the gain margin of identified system with the optimal controller and then multiply the controller gains by the gain margin which should result in marginally stable behavior.

Since this work is focused on vision-based control of the UAV where the delays are much significant and practically result in the systems where the optimality of the controller is not sensitive to parameter change hence a single class covers huge area in the parameter space, making it impossible to compare the system response for validation. 
\end{comment}
The verification of the adequacy of the identified time delay model is not feasible in experimentation, as the ground truth model is not available. However, it is possible to validate certain qualities of the identified time delay model experimentally, and compare such behavior with the delay-free model. One such quality is the stability limits of the system, i.e. the gain margin and the phase margin of the system. We chose to validate the model adequacy using the gain margin, as it is implementation is easier and less prone to inaccuracies introduced by digital discretization. 

\begin{figure} [t]
\centering
  \includegraphics[width = \linewidth]{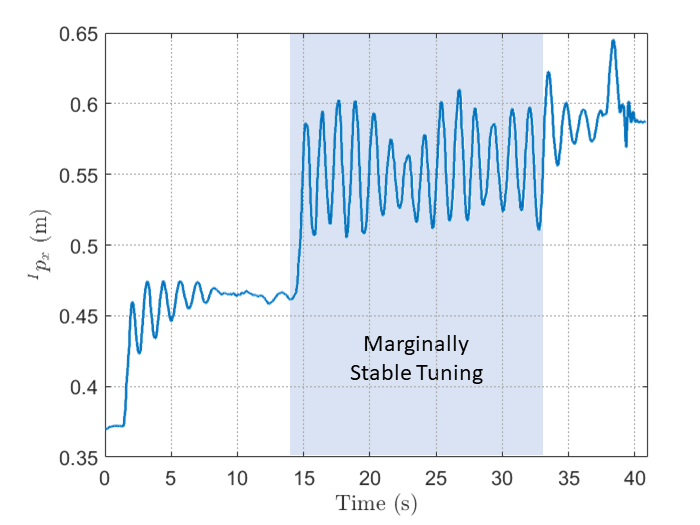}
\caption{Marginal stability response (highlighted) when PD controller gains are scaled by the calculated gain margin from identified system and controller parameters. }
\label{fig:marg_stable}
\end{figure}

We found the gain margin for the lateral loop along \(\bm{h_x}\) using the model parameters given in Table \ref{tab:id_cont} to be 1.4, hence we multiply the controller gains for the lateral loop along \(\bm{h_x}\) by 1.4 and a marginally stable behavior is observed as seen in Fig. \ref{fig:marg_stable}. For the case when the lateral loop delay is neglected, as widely considered in literature, the predicted gain margin would be close to 3.1, which is significantly different from the experimentally verified 1.4 value that is predicted by the time delay model. As such, it is clear that delay-free models cannot be used effectively for the design of UAV control systems.

\begin{figure} [htb]
\centering
  \includegraphics[width = \linewidth]{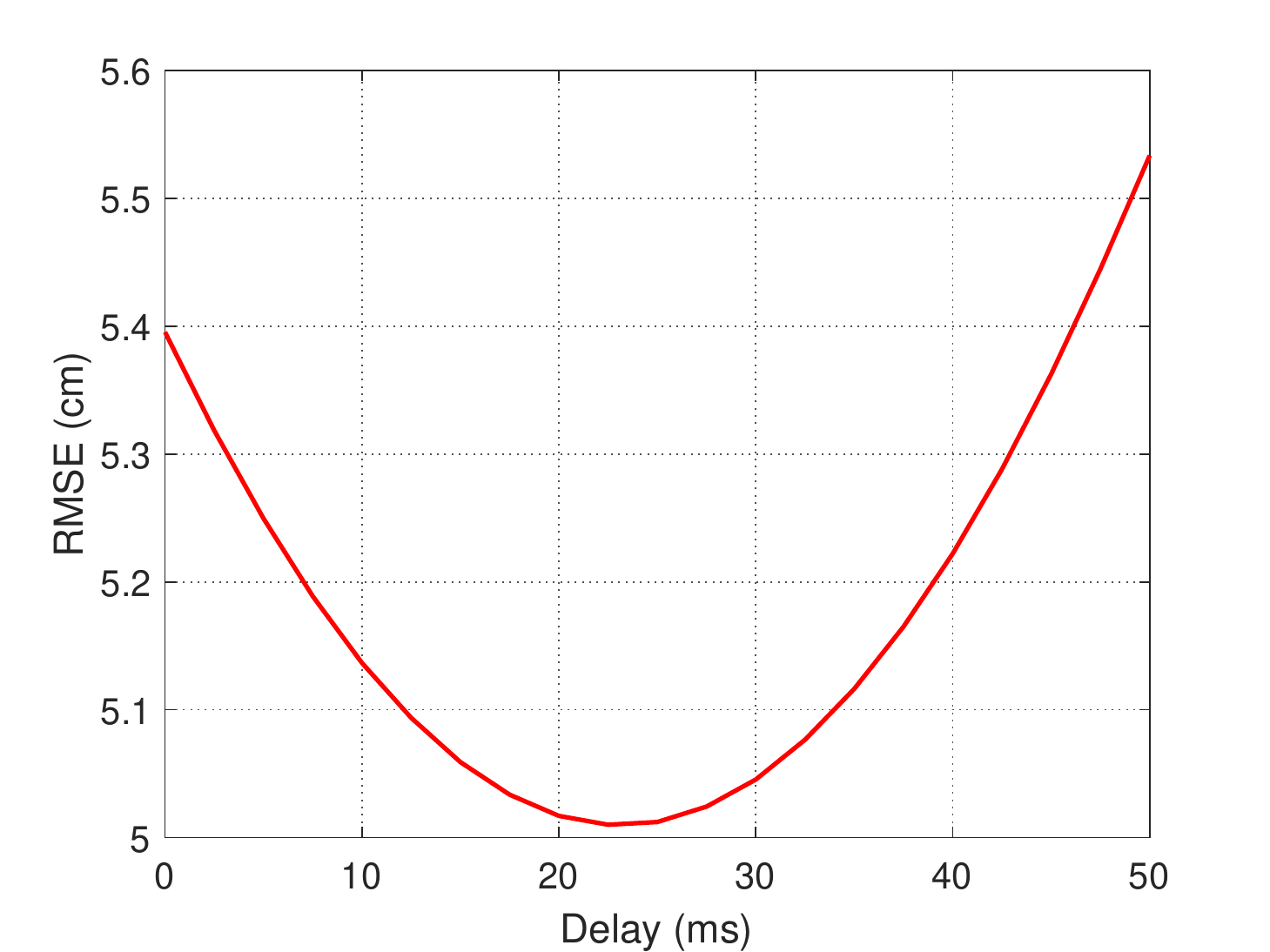}
\caption{RMSE with respect to time delay compensation. The compensation is simply performed by advancing the followed trajectory.Delay free models always predict an increase in RMSE when delay compensation is used.}
\label{fig:rmse_delay}
\end{figure}

In addition to the time delay model adequacy validation through stability limits, it is possible to further validate it by exploiting our knowledge of the time delay to implement delay compensation in trajectory generation to increase trajectory tracking performance. From equation \ref{eq_lateral_tf}, we see that there is a time delay in the numerator which corresponds to $\tau_\theta$ and entails that there will be a delay in tracking the reference. Hence, if we compensate for this delay, it should result in reduced RMSE. In Figure \ref{fig:rmse_delay}, we show the RMSE calculated for lemniscate trajectory experiment as explained in Section \ref{sec:traj_track} with different time delay compensations. We clearly see, there is a minimum at approximately 22 ms, which is very close to 17.5 and 20 ms delays we get from identification (refer to Table \ref{tab:id_cont}). Hence, it further validates the applicability of our analysis and approach experimentally.

\subsection{Trajectory Tracking Performance}
\label{sec:traj_track}
\begin{figure}[htb]
\centering
\includegraphics[width=\linewidth]{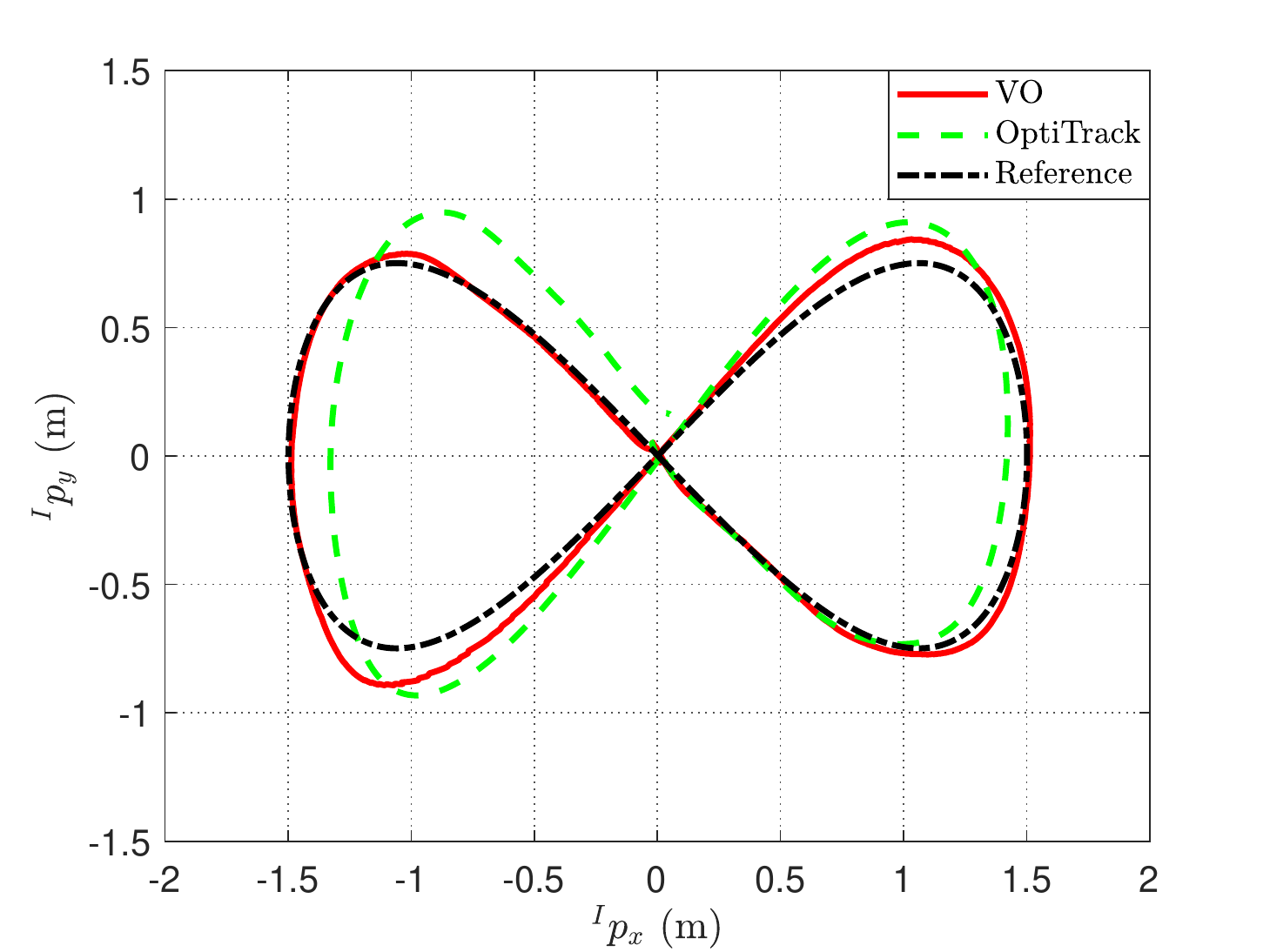}
\caption{Spatial trajectory tracking results in \(\bm{i_x}\times\bm{i_y}\) plane for lemniscate trajectory using vision-based control. OptiTrack provides the actual position of the UAV which shows how vision-based position estimation (VO) drifts over time. }
\label{fig:ref_track}
\end{figure}

\begin{figure}[t]
\centering
\includegraphics[width=\linewidth]{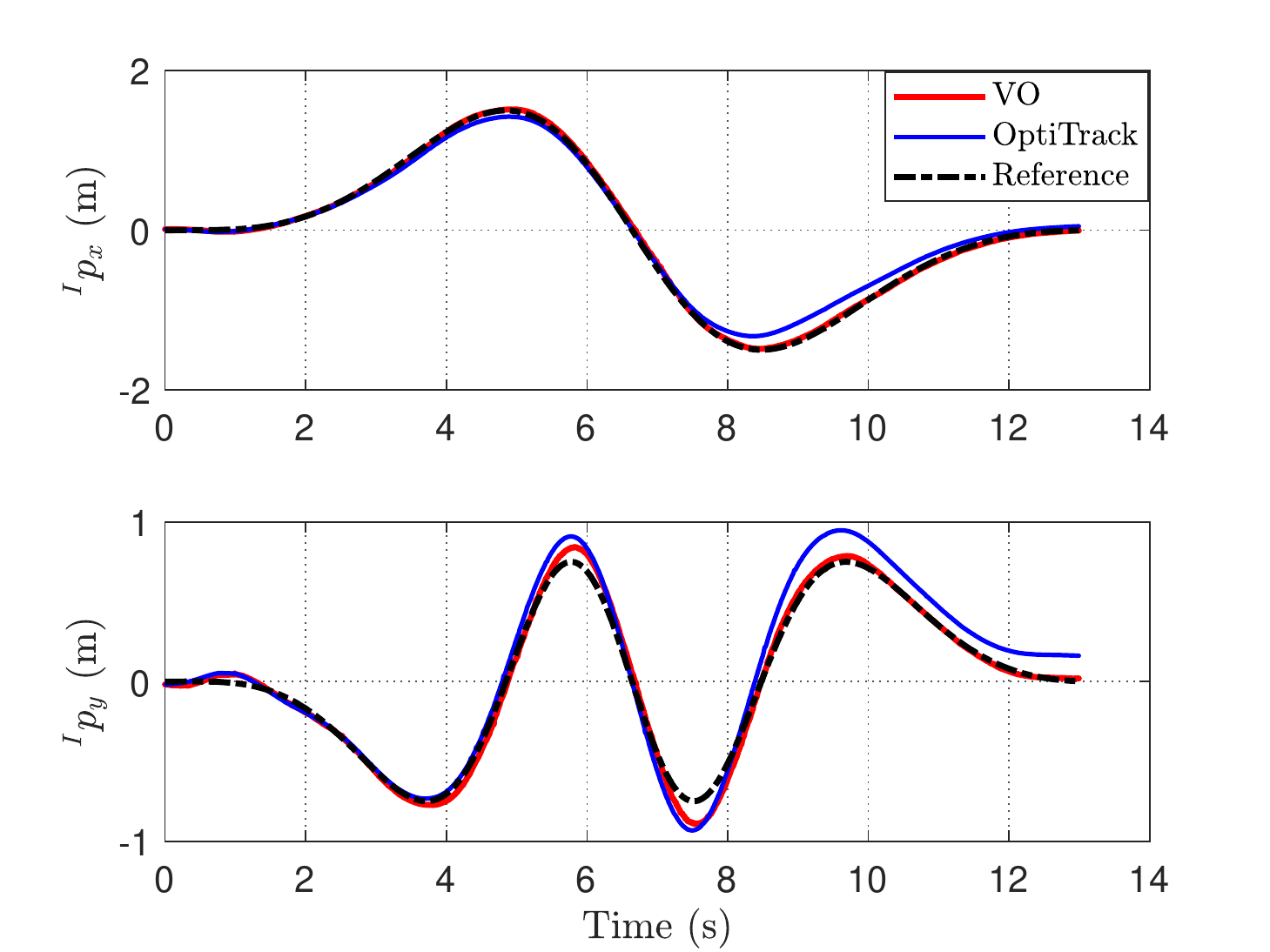}
\caption{Reference tracking for lateral loops during lemniscate trajectory with on-board vision-based control. OptiTrack is also provided to show estimation errors in vision-based position estimation (VO).}
\label{fig:2d_track}
\end{figure}

The best way to evaluate controller performance in this case would be for trajectory tracking. Due to added delays and estimation uncertainties, controller tuning is a very difficult task for vision-based control of UAVs. To gain stability, too conservative controllers will result in poor trajectory performance. To the best of our knowledge, most of the works in vision-based control of UAV report poor trajectory tracking performance. The best reported results we have found are in \cite{loianno2016estimation} and \cite{shen2013vision}, but they do not provide quantitative results for control RMSE. Hence, comparing qualitatively, we get the trajectory tracking performance on par if not exceeding as shown in Figures \ref{fig:ref_track} and \ref{fig:2d_track}. We are flying a planar trajectory, hence we only show the results for \(\bm{i_x}\times\bm{i_y}\) plane for better visualization, however the UAV is using vision-based control for all translational loops. We chose lemniscate trajectory since it is a fairly complex trajectory and widely used in the literature for performance evaluation. We generate a minimum snap trajectory using the method described in \cite{mellinger2011minimum} to traverse $3 \times 1.5$ meter lemniscate in 13 seconds.

To assess the vision based trajectory tracking performance, we use two metrics, the first one is the root mean squared error (RMSE) which is given by:
\begin{equation}
\label{eq:rmse}
    RMSE = \sqrt{\dfrac{\sum _{i=1} ^N ({}^Ip_x-{}^Ip_x^{ref})_i^2 + ({}^Ip_y-{}^Ip_y^{ref})_i^2}{N}}
\end{equation}
where $N$ is the number of the sampled data points in the trajectory. RMSE provides a single measurement for evaluating the temporal tracking of reference in x and y both. The second metric is contouring error (CE), which captures how good the spatial tracking is, and is defined by the minimum Euclidean distance between a given position measurement and the nearest point on the reference trajectory:

\begin{gather}
    CE_i = \min \left( \sqrt{({}^Ip_{x_i} - {}^Ip_x^{ref})^2 + ({}^Ip_{y_i} - {}^Ip_y^{ref})^2} \right)\\
    \label{eq:CEmax}
    CE_{max} = \max \left( CE_1, ... , CE_N \right)\\
    \label{eq:CE}
    CE = \frac{\sum_i^N CE_i}{N}
\end{gather}

In Table \ref{tab:traj_results}, we provide the quantitative evaluation of our trajectory tracking results, showing the metrics for spatial tracking (CE) and temporal tracking (RMSE) performances. We also provide estimation RMSE, since it also affects the control performance. These results can be used as benchmark for future works on vision-based trajectory tracking with UAVs. Maximum velocity has also been reported here, since tracking error also depends on the velocity.

\begin{table}[t]
    \centering
    \begin{tabular}{|c||c|c|c|c|c|}
        \hline
         Trajectory & $CE_{max}$ & CE & $V_{max}$ & $RMSE$ & $RMSE_{est}$ \\
         \hline
         lemniscate & 0.147 & 0.027 & 2.09 & 0.054 & 0.136 \\
         \hline
    \end{tabular}
    \caption{RMSE, $CE_{max}$ and CE are defined in eq \ref{eq:rmse}, \ref{eq:CEmax} and \ref{eq:CE}, respectively. While $V_{max}$ is maximum velocity during the trajectory and $RMSE_{est}$ is the root mean squared error between VO and motion capture. All quantities are in SI units.}
    \label{tab:traj_results}
\end{table}

\section{Conclusion}
This paper investigated the use of nonlinear time delay models to describe multirotor UAV dynamics with applications to vision based navigation. It was demonstrated that the proposed modeling framework has the advantage of being suitable for controller tuning compared with the widely used delay free models. Introducing time delay changes the stability behavior of the system. It was found that the delay free models has a globally minimum feedback cost that happens when controller gains are infinite, which is not the case with time delay models. The adequacy of the time delay model was demonstrated experimentally, as the model accurately predicted the stability limits of the system and, it also predicted an increase in trajectory tracking performance due to delay compensation. The model parameters are obtained systematically using DNN-MRFT and, without manual tuning or refinement, the designed controller was able to achieve trajectory tracking performance that is on par with the state-of-the-art.

In a future work, we will investigate the possible use of feedforward terms to improve trajectory tracking performance based on the linearized models presented in Section \ref{sec_time_delay_model}. Also, the obtained models can be used to efficiently plan for trajectories while accounting for states and inputs constraints.

%%%%%%%%%%%%%%%%%%%%%%%%%%%%%%%%%%%%%%%%%%%%%%%%%%%%%%%%%%%%%%%%%%%%%%%%%%%%%%%%

%%%%%%%%%%%%%%%%%%%%%%%%%%%%%%%%%%%%%%%%%%%%%%%%%%%%%%%%%%%%%%%%%%%%%%%%%%%%%%%%

%%%%%%%%%%%%%%%%%%%%%%%%%%%%%%%%%%%%%%%%%%%%%%%%%%%%%%%%%%%%%%%%%%%%%%%%%%%%%%%%

\section*{ACKNOWLEDGMENT}

We would like to thank Eng. Abdulla Ayyad for helping with DNN-MRFT. We also thank Eng. Oussama AbdulHay and Romeo Sumeracruz for their help in preparing the experimental setup.

%%%%%%%%%%%%%%%%%%%%%%%%%%%%%%%%%%%%%%%%%%%%%%%%%%%%%%%%%%%%%%%%%%%%%%%%%%%%%%%%

\bibliographystyle{IEEEtran}

\bibliography{bst/ref_items.bib}

\begin{IEEEbiography}[{\includegraphics[width=1in,height=1.25in,clip,keepaspectratio]{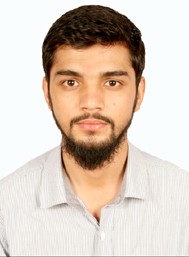}}]%
{Muhammad Ahmed Humais}
received his M.Sc. in Electrical and Computer Engineering from Khalifa University in 2020. His research is mainly focused on robotic perception and control for autonomous systems. He is currently a Ph.D. fellow at Khalifa University Center for Autonomous Robotics (KUCARS). 
\end{IEEEbiography}

\begin{IEEEbiography}[{\includegraphics[width=1in,height=1.25in,clip,keepaspectratio]{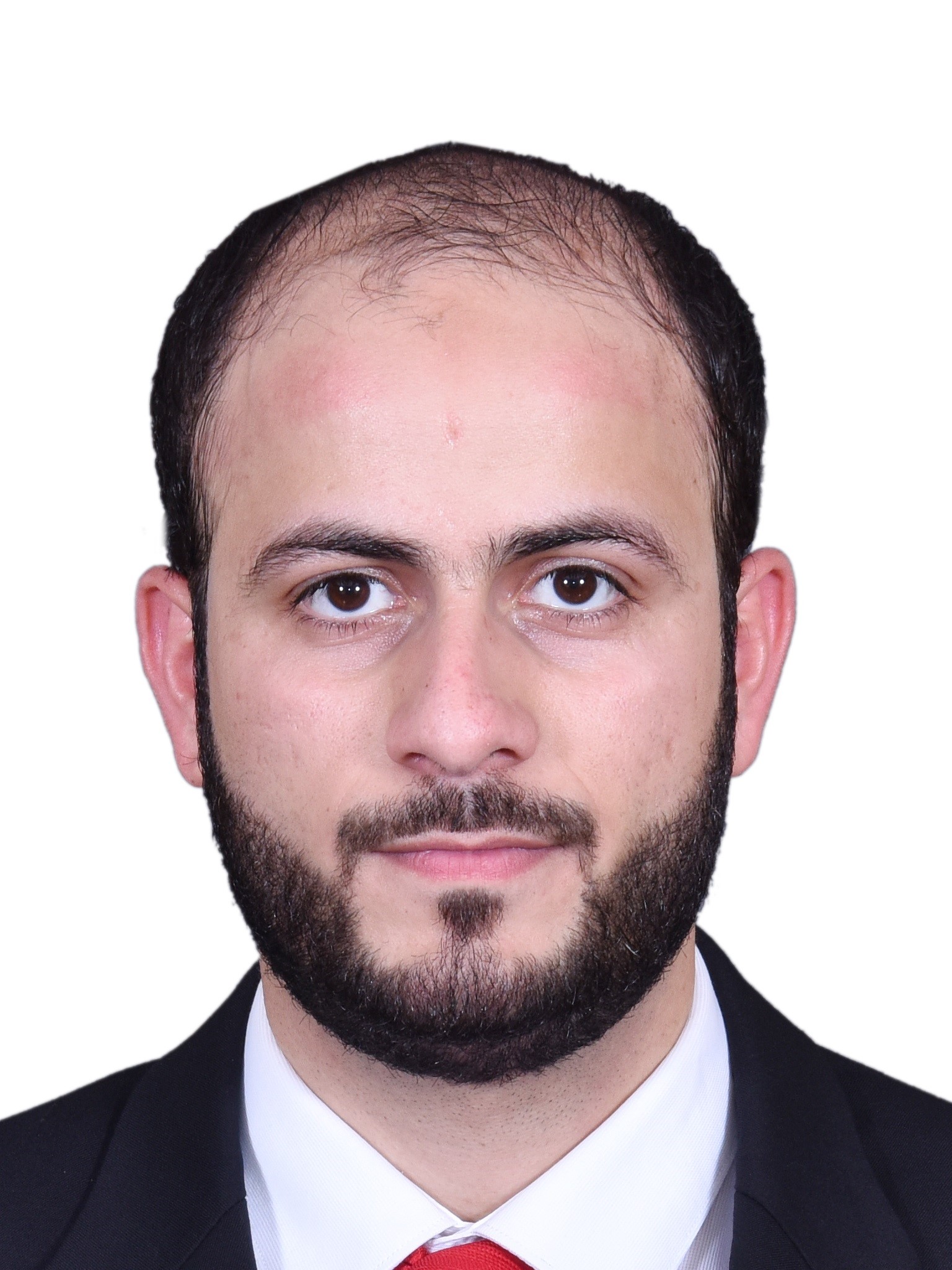}}]{Mohamad Chehadeh} received his MSc. in Electrical Engineering from Khalifa University, Abu Dhabi, UAE, in 2017. He is currently with Khalifa University Center for Autonomous Robotic Systems (KUCARS). His research interest is mainly focused on identification, perception, and control of complex dynamical systems utilizing the recent advancements in the field of AI.
\end{IEEEbiography}

\begin{IEEEbiography}[{\includegraphics[width=1in,height=1.25in,clip,keepaspectratio]{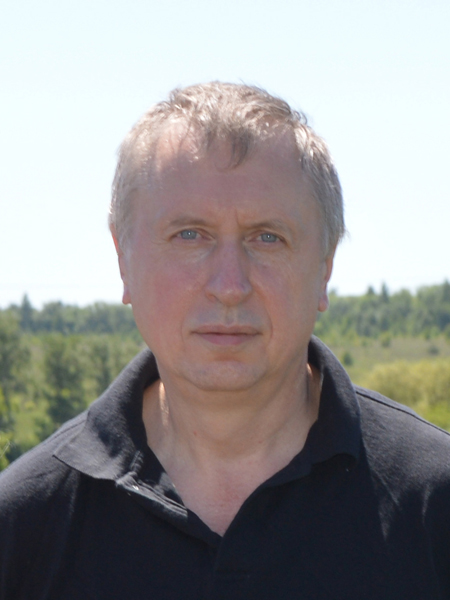}}]{Igor Boiko}
received his MSc, PhD and DSc degrees from Tula State University and Higher Attestation Commission, Russia. His research interests include frequency-domain methods of analysis and design of nonlinear systems, discontinuous and sliding mode control systems, PID control, process control theory and applications. Currently he is a Professor with Khalifa University, Abu Dhabi, UAE.
\end{IEEEbiography}

\begin{IEEEbiography}[{\includegraphics[width=1in,height=1.25in,clip,keepaspectratio]{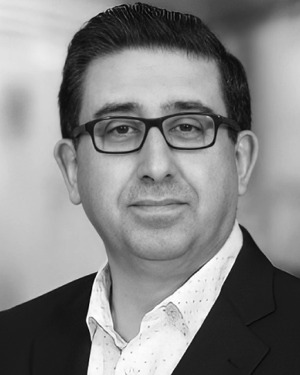}}]{Yahya Zweiri} received the Ph.D. degree from the King’s College London in 2003. He is currently an Associate Professor with the Department of Aerospace Engineering and deputy director of Advanced Research and Innovation Center - Khalifa University, United Arab Emirates. He was involved in defense and security research projects in the last 20 years at the Defense Science and Technology Laboratory, King’s College London, and the King Abdullah II Design and Development Bureau, Jordan. He has published over 110 refereed journals and conference papers and filed ten patents in USA and U.K., in the unmanned systems field. His main expertise and research are in the area of robotic systems for extreme conditions with particular emphasis on applied Artificial Intelligence (AI) aspects and neuromorphic vision system.
\end{IEEEbiography}


\end{document}